\documentclass[sigconf, table, dvipsnames]{acmart}
\usepackage{listings}
\usepackage{booktabs}
\usepackage{array}
\usepackage{multirow}
\usepackage{enumitem}
\usepackage[linesnumbered,ruled,vlined]{algorithm2e}
\usepackage{lineno}
\usepackage{subcaption}
\DeclareUnicodeCharacter{2264}{\ensuremath{\leq}}

\renewcommand\footnotetextcopyrightpermission[1]{} 

\AtBeginDocument{%
  \providecommand\BibTeX{{%
    \normalfont B\kern-0.5em{\scshape i\kern-0.25em b}\kern-0.8em\TeX}}}

\graphicspath{{./images/}} 
\begin{document}

\title{Uncertainty-Aware Multimodal Anti-UAV Detection via Evidential Fusion and Conflict-Discounted Belief Aggregation}

\author{Sharanda Suttorp}
\affiliation{%
  \institution{University of Amsterdam}
  \city{Amsterdam}
  \country{The Netherlands}
}
\email{sharanda.suttorp@student.uva.nl}

\author{Seyed Sahand Mohammadi Ziabari}
\authornote{Corresponding author.}
\affiliation{%
  \institution{University of Amsterdam}
  \city{Amsterdam}
  \country{The Netherlands}
}
\affiliation{%
  \institution{Department of Computer Science and Technology, SUNY Empire State University}
  \city{Saratoga Springs}
  \state{NY}
  \country{USA}
}
\email{s.s.mohammadiziabari@uva.nl}

\author{Ali Mohammed Mansoor Alsahag}
\affiliation{%
  \institution{University of Amsterdam}
  \city{Amsterdam}
  \country{The Netherlands}
}
\email{a.m.m.alsahag@uva.nl}

\begin{abstract}

{
Anti-UAV perception systems must remain reliable when sensor streams degrade under occlusion, fast motion, or modality-specific failure. Existing multimodal anti-UAV systems fuse RGB and thermal streams deterministically, without modeling predictive uncertainty, and cannot express doubt when streams disagree. Evidential Deep Learning (EDL) produces calibrated per-class uncertainty in a single forward pass. EDTC already exploits this for thermal-only perception, yet cross-modal evidential fusion remains unaddressed. This paper extends EDTC to multimodal RGB-Thermal perception via Discounted Belief Fusion (DBF), which converts inter-modal conflict into uncertainty mass before aggregating stream opinions. Bounding boxes are resolved by selecting the lower-uncertainty modality. On the Anti-UAV benchmark, multimodal fusion consistently outperforms either single stream (test Acc 0.670 vs. 0.604 IR, 0.598 RGB) at real-time speed ($\ge$38 FPS). However, DBF is empirically indistinguishable from undiscounted averaging: near-zero inter-modal conflict on this presence-dominated benchmark leaves the discounting step inert. The fused uncertainty is well-calibrated (ECE 0.057) yet expectedly a weaker localization failure detector than spatial variance (AUROC 0.626 vs. 0.739). The null result is structural: the benchmark's near-universal presence and vacuous miss-encoding jointly suppress inter-modal conflict, a diagnosis that delimits where conflict-aware fusion provides measurable benefit.}

\end{abstract}

\keywords{Anti-UAV Perception, Evidential Deep Learning, Multimodal Fusion, Discounted Belief Fusion} 


\fancyhead{}
\maketitle
\section{Introduction}
\label{ch:Intro}

As Unmanned Aerial Vehicles (UAVs) are becoming more prevalent, the need for reliable airspace security has become increasingly urgent. While drones have many beneficial applications, unauthorized UAV activity poses risks to public safety and privacy. Therefore, UAV detection and tracking systems are essential to guarantee airspace safety, placing this work in the intersection of computer vision and safety-critical perception.

In real world anti-UAV settings, sensor data is inherently noisy and degrades under adverse environmental conditions. Challenges such as fast motion, occlusion, scale variation and temporary target disappearance significantly complicate reliable detection and tracking \cite{Katona2025MARINE}. To reflect these practical difficulties, recent benchmarks such as Anti-UAV \cite{Jiang2023ANTIUAV} and AntiUAV600 \cite{Zhu2023EDTC} have introduced large-scale datasets featuring diverse and challenging scenarios. Notably, AntiUAV600 formulates anti-UAV perception as a continuous detection and tracking problem without assuming prior target initialization. This reflects real-world settings more closely.

Despite these advances, most existing approaches rely on single-modality frameworks \cite{Zhu2023EDTC, zhao2022visionbasedantiuavdetectiontracking, Jiang2023ANTIUAV} or deterministic fusion strategies \cite{larrat2025multimodaltransformerapproachuav, zongzhen2025crossmodaloffsetguideddynamicalignment, alla_trident_2025}. Multimodal fusion offers a way to improve robustness by leveraging complementary sensor streams. Yet, current multimodal fusion strategies for anti-UAV perceptions, such as the cross-attention transformer of Larrat et al. \cite{larrat2025multimodaltransformerapproachuav} and the cross-modal alignment of Zongzhen et al. \cite{zongzhen2025crossmodaloffsetguideddynamicalignment}, fuse modality-specific features without producing any estimate of predictive uncertainty. Critically, this means that when streams disagree (RGB degrading at night, thermal failing under thermal crossover) the system has no principled way to express doubt or defer. As a result, these systems may commit confidently to incorrect predictions precisely under the conditions where caution is the most needed.

Evidential Deep Learning (EDL) \cite{sensoy2018evidentialdeeplearningquantify, Gao2025EDL} offers a promising direction: it produces calibrated, per-class uncertainty in a single forward pass via Dirichlet distributions over class probabilities, making it suitable for real-time deployment. The Evidential Detection and Tracking Collaboration (EDTC) framework \cite{Zhu2023EDTC} already exploits EDL in this way, using an evidential head on its tracking branch to gate switching between global detection and local tracking. However, EDTC operates on thermal infrared alone and does not address the fusion of conflicting evidence across modalities. When evidential opinions from independent RGB and thermal streams must be combined, the choice of fusion rule matters: Average Belief Fusion (ABF) \cite{Josang2017SL} correctly handles dependent sources, but recent work by Bezirganyan et al. \cite{bezirganyan2025multimodallearninguncertaintyquantification} shows it underestimates global uncertainty when streams are both highly confident but contradictory. Whether and how conflict-aware evidential fusion can address this limitation in the anti-UAV setting remains an open question.

This paper investigates that question by extending EDTC from single-modality thermal to multimodal RGB-Thermal perception via Discounted Belief Fusion (DBF) \cite{bezirganyan2025multimodallearninguncertaintyquantification} which discounts conflicting beliefs into uncertainty before averaging. Per-modality spatial uncertainty is captured via Gaussian bounding box heads \cite{he2019boundingboxregressionuncertainty} with bounding boxes resolved at each frame by selecting the lower-uncertainty modality (Dynamic Modality Selection). Whether genuine inter-modal conflict arises under the properties of the Anti-UAV benchmark, and whether DBF's discounting mechanism can therefore differentiate itself from undiscounted averaging, is itself an empirical question that this research sets out to answer.

This study investigates the impact of uncertainty-aware multimodal fusion on the predictive performance, uncertainty calibration, robustness, and computational efficiency of real-time anti-UAV detection. The evaluation focuses on the impact of conflict-aware belief fusion on detection performance relative to undiscounted, non-uncertainty-aware, and single-stream baselines, as measured by AP@0.5, mAP, mIoU, and Acc. The calibration of the uncertainty estimates produced by the proposed fusion framework is assessed using ECE, while AUROC is used to evaluate whether the fused uncertainty can reliably discriminate between correct and incorrect predictions. The robustness of uncertainty-aware multimodal fusion is further examined under challenging anti-UAV conditions, including occlusion, fast motion, and thermal crossover, with comparisons against single-modality baselines for each attribute subset. Finally, the computational efficiency of the dual-stream uncertainty-aware architecture is evaluated against single-modality baselines to determine whether it meets real-time constraints and how the choice of fusion operator affects throughput.

By systematically evaluating these aspects on the Anti-UAV benchmark, this research aims to deliver both a working multimodal extension of EDTC and an empirical account of when conflict-aware evidential fusion provides measurable benefit over simpler fusion strategies.


\section{Anti-UAV Detection, Tracking, and Benchmarks}

Vision-based anti-UAV perception is challenging because drones are typically
small, fast, and easily confused with clutter such as birds, buildings, and
clouds. Small-object detection and tracking are particularly challenging under
occlusion and degraded visual conditions, where limited target information can
reduce localization and tracking reliability
\cite{AnwarZiabari2025,dong2025securingskiescomprehensivesurvey}. For the tracking task it was usually assumed that the UAV appears in the initial frame and then followed over time. However, in real deployments, anti-UAV systems require to do both tasks of detection and tracking. A system must discover a drone when it enters the scene, track it while it is visible, and then re-detect it when it has gone out of view for a certain time period \cite{Zhu2023EDTC, dong2025securingskiescomprehensivesurvey}.

Recent key benchmarks include DUT Anti-UAV \cite{zhao2022visionbasedantiuavdetectiontracking} which offers an RGB dataset of a detection set (10k images) and a tracking set (20 videos), explicitly encouraging detector-tracker combinations. Anti-UAV \cite{Jiang2023ANTIUAV} is a large RGB-Thermal benchmark with 318 paired RGB-Thermal sequences and 580k annotated boxes. AntiUAV600 + EDTC \cite{Zhu2023EDTC} targets realistic surveillance without prior target initialization: 600 thermal sequences, 723k frames, and the EDTC baseline that uses evidential uncertainty to gate detection-tracking switching. MMAUD \cite{YuanMMAUD} expands beyond vision with synchronized cameras, LiDAR, mmWave radar and audio. MMAUD reflects the field's move toward multi-sensor anti-drone perception. Similarly, TRIDENT \cite{alla_trident_2025} proposes a tri-modal (audio, video and radio frequency) real-time drone detection framework and dataset. Together, these benchmarks establish the evaluation protocol and challenge-attribute subsets used in this study to systematically assess fusion performance and robustness under varying anti-UAV operating conditions.



\subsection{Multimodal Fusion}

Multimodal learning combines complementary sensors to improve robustness when any single stream degrades. Baltrušaitis et al. \cite{baltrušaitis2017multimodalmachinelearningsurvey} provide a foundational taxonomy organizing fusion around representation, alignment and integration. Jiao et al. \cite{JIAO20241} categorize it by pipeline stage: early fusion (data-level), deep fusion (feature-level), late fusion (decision-level) and hybrid setups that combine multiple stages. With the core trade-off between cross-modal information sharing and robustness to missing or noisy streams.


For anti-UAV perception specifically, Larrat et al. \cite{larrat2025multimodaltransformerapproachuav} fuse radar, audio and video via a cross-attention transformer for UAV detection. Zongzhen et al. \cite{zongzhen2025crossmodaloffsetguideddynamicalignment} propose cross-modal offset-guided dynamic alignment for weakly-aligned RGB-Thermal UAV detection. TRIDENT \cite{alla_trident_2025} evaluates late and gated fusion under diverse conditions. A shared limitation across these works is that when streams disagree none has a principled mechanism to express doubt or defer. All fuse without modeling predictive uncertainty. This motivates the evaluation of whether uncertainty-aware fusion produces well-calibrated uncertainty estimates that reliably discriminate between correct and incorrect predictions.


Bezirganyan et al. \cite{bezirganyan2025multimodallearninguncertaintyquantification} address this directly: they show that Averaging Belief Fusion (ABF) \cite{Josang2017SL}, which handles dependent sources correctly under Subjective Logic, underestimates global uncertainty when streams are confident but contradictory, and propose Discounted Belief Fusion (DBF) as a remedy. DBF introduces a conflict-based discounting step before averaging, converting cross-modal conflict into additional uncertainty mass. The Dempster-Shafer framework \cite{dempster, fusionsubjectivelogic} underlying both operators is inappropriate in its original form under source dependence, making ABF the correct undiscounted baseline rather than Dempster's rule. This paper adopts DBF for classification fusion. We hypothesize that DBF will outperform ABF on frames where the streams are confidently contradictory. Whether such conflicts arise within the Anti-UAV benchmark is evaluated empirically through the comparative fusion analysis.


\subsection{Evidential Deep Learning}
\label{sec:edl_background}

Quantifying predictive uncertainty in deep networks is commonly approached through Bayesian methods such as MC Dropout \cite{gal2016dropoutbayesianapproximationrepresenting} or deep ensembles \cite{lakshminarayanan2017simplescalablepredictiveuncertainty}, which estimate uncertainty by aggregating multiple stochastic forward passes. While effective, these methods incur a linear computational overhead in the number of passes, making them poorly suited to real-time applications with strict latency constraints. Evidential Deep Learning (EDL) addresses this by producing both class predictions and calibrated uncertainty in a single deterministic forward pass \cite{Gao2025EDL}. This property makes EDL particularly attractive for real-time anti-UAV perception. Yet whether this efficiency advantage is retained when extending the model to a dual-stream architecture with an additional fusion operator remains an open empirical question and is therefore evaluated in this study.


Instead of outputting softmax probabilities, an EDL model outputs non-negative evidence $e_k \geq 0$ for each class $k$ via a ReLU activation. These are mapped to Dirichlet concentration parameters $\alpha_k = e_k + 1$, parameterizing a Dirichlet distribution over class probabilities. The total Dirichlet strength $S = \sum_{k=1}^{K} \alpha_k$ controls the concentration of this distribution: large $S$ indicates confident predictions, small $S$ indicates high uncertainty \cite{sensoy2018evidentialdeeplearningquantify}.

EDL is motivated by Subjective Logic in which predictions are represented as opinions consisting of per-class belief masses $b_k$ and an explicit uncertainty mass $u$ \cite{fusionsubjectivelogic, sensoy2018evidentialdeeplearningquantify}:
\[
b_k = \frac{e_k}{S}, \qquad u = \frac{K}{S}, \qquad \sum_{k=1}^{K} b_k + u = 1.
\]

Sensoy et al. \cite{sensoy2018evidentialdeeplearningquantify} propose several loss variants for training the evidential head. This work adopts the expected cross-entropy form, which is the most widely used variant and the one employed by EDTC. This loss combines an expected cross-entropy term that encourages evidence for the correct class with a KL regularization term that penalizes spurious evidence on incorrect classes:
\[
\mathcal{L}_{EDL} = \underbrace{\sum_{k=1}^{K} y_k \left(\log S - \log \alpha_k\right)}_{\text{expected cross-entropy}} + \; \lambda_{KL} \cdot \underbrace{\mathrm{KL}\!\left[\mathrm{Dir}(\mathbf{p} \mid \tilde{\boldsymbol{\alpha}}) \;\|\; \mathrm{Dir}(\mathbf{p} \mid \mathbf{1})\right]}_{\text{regularization toward uniform prior}}
\]
where $\tilde{\alpha}_k = y_k + (1 - y_k) \cdot \alpha_k$ removes the evidence for the ground-truth class before computing the KL divergence, ensuring the regularizer only penalizes evidence on incorrect classes. The weight $\lambda_{KL}$ is typically annealed from 0 to 1 during early training to allow the network to first learn discriminative evidence before the regularizer prevents overconfidence \cite{sensoy2018evidentialdeeplearningquantify}.

EDTC applies EDL to the tracking branch only and provides no mechanism for fusing conflicting evidential opinions across modalities. This paper addresses this gap by evaluating conflict-aware multimodal fusion and the calibration of the resulting uncertainty estimates.

\section{Methodology}
\label{sec:methodology}



\subsection{Description of the Data}
\label{sec:method-data}
Experiments use the Anti-UAV dataset \cite{Jiang2023ANTIUAV}: 318 paired RGB-Thermal sequences with over 580k annotated bounding boxes, split into training (160), validation (67), and test (91). Each frame includes bounding boxes, a visibility flag, and seven challenge-attribute labels: thermal crossover (TC), fast motion (FM), scale variation (SV), low illumination (LI), low resolution (LR), out-of-view (OV) and occlusion (OC). Official splits are followed throughout.

\begin{figure}[htbp]
    \centering
    \includegraphics[width=0.49\columnwidth]{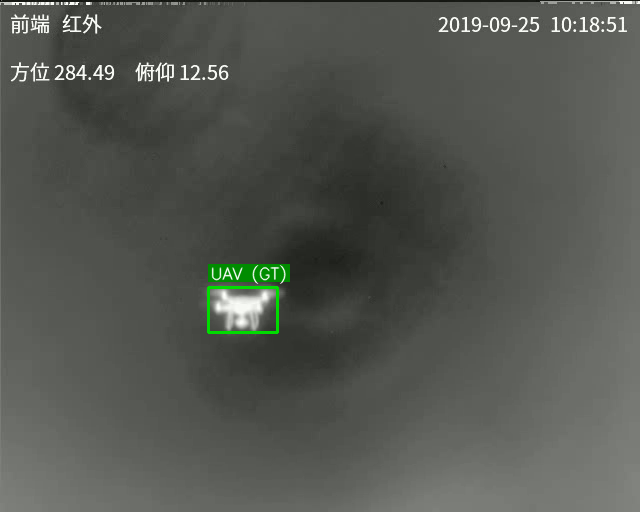}
    \hfill
    \includegraphics[width=0.49\columnwidth]{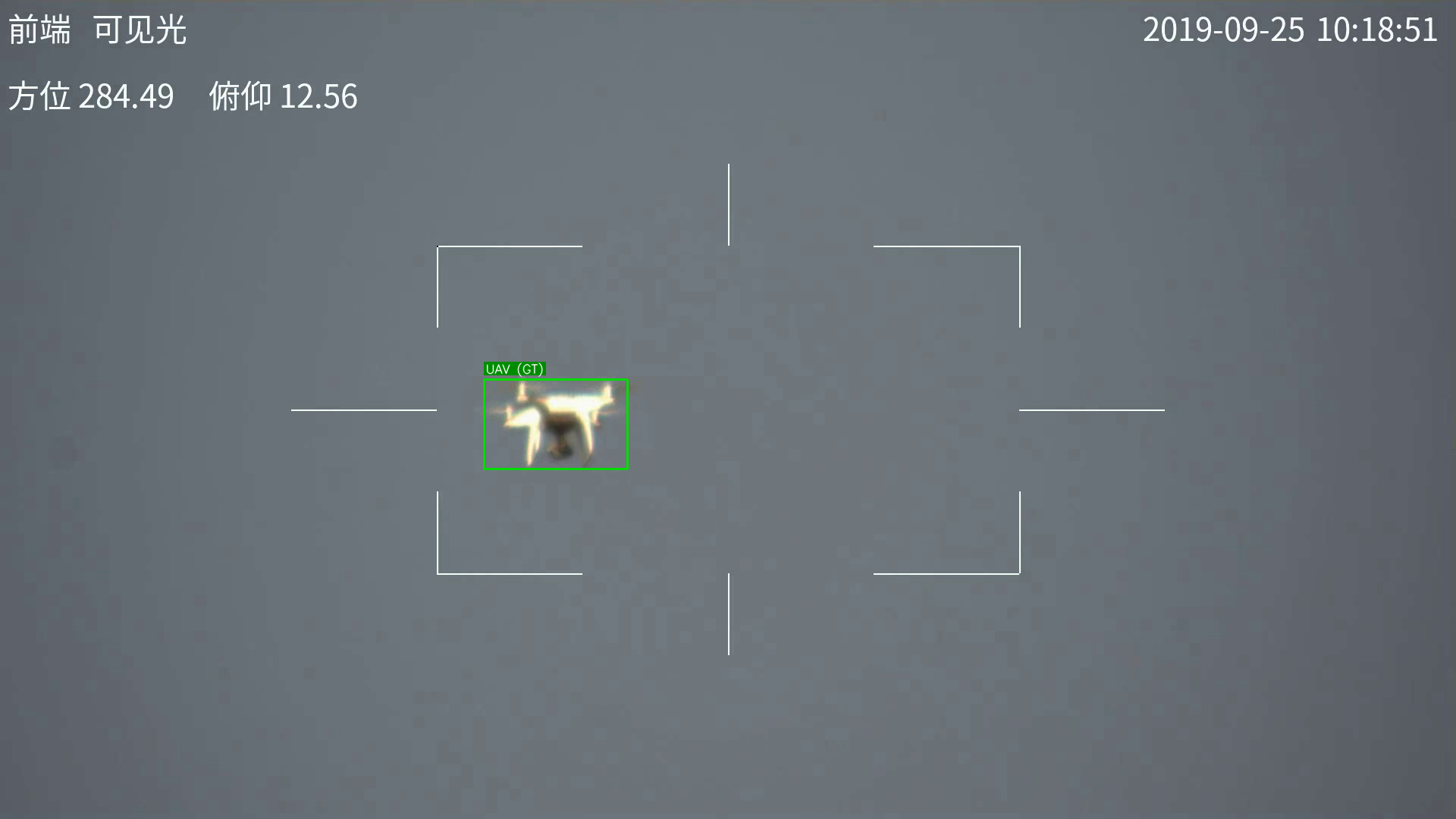}
    \caption{Representative paired frames from the Anti-UAV dataset with ground-truth bounding boxes (green). The infrared (left) and RGB (right) streams differ in resolution ($640\!\times\!512$ vs. $1920\!\times\!1080$) and are spatially unaligned, making direct cross-modal bounding box fusion ill-posed.}
    \label{fig:gt_paired}
\end{figure}

Exploratory analysis reveals properties that directly inform modeling choices. UAVs appear as small targets with median normalized scales of $0.060$ (visible) and $0.065$ (infrared). Absolute box sizes differ markedly: median visible box $116\!\times\!61$\,px (frame $1920\!\times\!1080$) vs.\ median infrared box $48\!\times\!28$\,px (frame $640\!\times\!512$). This asymmetry makes direct cross-modal bounding box fusion ill-posed and motivates the independent spatial tracking design (Section~\ref{sec:edtc_switching}). Figure~\ref{fig:gt_paired} illustrates this asymmetry on a representative frame pair. The modalities are complementary but infrared-dominant: visibility labels agree in 95.1\% of frames, but when they diverge, infrared detects the drone while visible does not in 4.6\% of cases, whereas the reverse occurs in only 0.26\%. The dataset is strongly presence-dominated, with the UAV present in ${\sim}94\%$ of visible and ${\sim}99\%$ of infrared frames across all splits. Temporally, the median continuous presence segment spans 899~frames while absence segments last a median of only 28~frames, making the task predominantly one of continuous tracking with brief interruptions. Among sequence-level challenge attributes, thermal crossover is the most prevalent (60\% of sequences), followed by fast motion (38\%) and low illumination (37\%).

\subsection{Framework Design}
\label{sec:method-framework}
This research extends the EDTC framework \cite{Zhu2023EDTC} from single-modality (thermal) to multimodal RGB-Thermal anti-UAV perception. The core contribution is an uncertainty-aware fusion mechanism operating on per-stream evidential outputs. 

\begin{figure}[htbp]
    \centering
    \includegraphics[width=1\linewidth]{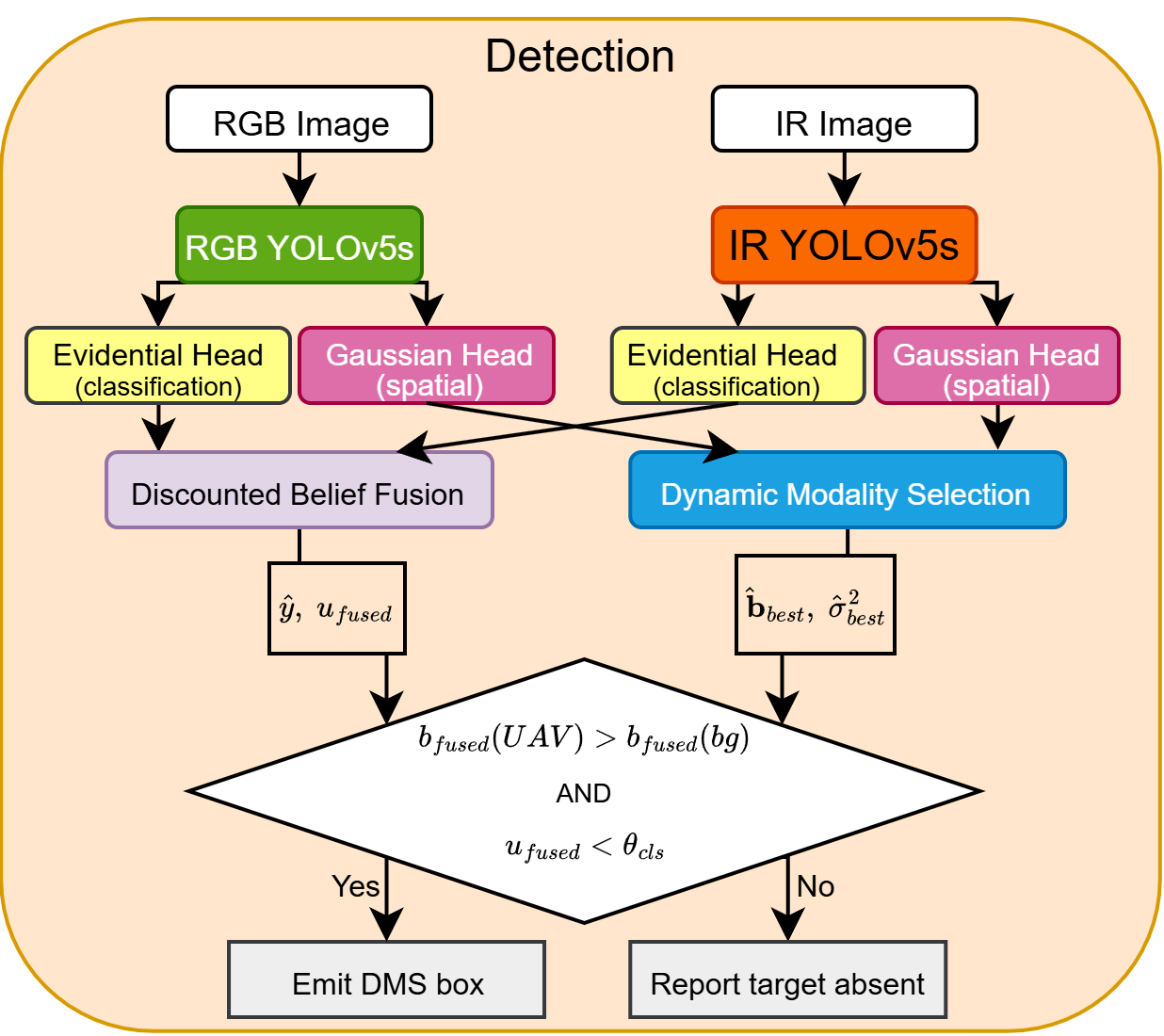}
    \caption{Multimodal uncertainty-aware detection framework. The detection branch (YOLOv5s) processes RGB and infrared streams independently, each producing per-modality semantic uncertainty via an evidential head and spatial uncertainty via a Gaussian head. Semantic opinions are aggregated across modalities via Discounted Belief Fusion (DBF), and the bounding box is resolved by selecting the lower-spatial-uncertainty modality (Dynamic Modality Selection). The fused opinion is passed through an uncertainty-gated output decision: a box is emitted only when the fused belief favors the UAV class ($b_\text{fused}(\text{UAV}) > b_\text{fused}(\text{bg})$) and the fused uncertainty falls below the gate threshold ($u_\text{fused} < \theta_\text{cls}$); otherwise the target is reported absent.}
    \label{fig:architecture}
\end{figure}

The framework was designed and implemented as a full detect-track hybrid following EDTC, with a SiamCAR-based \cite{guo2019siamcarsiamesefullyconvolutional} local tracking branch carrying the same evidential and Gaussian heads (see Appendix~\ref{app:det-track} for the full architecture and preliminary results). However, preliminary validation showed the detection-only configuration matching or exceeding the detect-track variant within each modality on EDTC Acc. All main results therefore use the detection-only configuration, which retains the uncertainty-gated commit decision of EDTC's switching logic but applies it as an output gate rather than a state transition. The remainder of this section describes the detection-only configuration. Figure~\ref{fig:architecture} provides a high-level overview of the architecture.

\subsubsection{Dual-stream architecture with uncertainty heads}
\label{sec:architecture}

The framework processes RGB and thermal inputs through parallel, independent streams. No parameters are shared between modalities, preventing a degraded stream from corrupting the other's representations. The global detection branch uses YOLOv5s \cite{redmon2016lookonceunifiedrealtime}, matching EDTC. 

Each stream is equipped with an evidential head (linear layer + ReLU) outputting $K{=}2$ non-negative evidence values interpreted as Dirichlet parameters. In the detection branch, this extends YOLOv5s with evidential classification (EDTC's original detector had no uncertainty estimation). 
Additionally, each stream is augmented with a Gaussian head \cite{he2019boundingboxregressionuncertainty} predicting mean $\mu$ and log-variance $\log(\sigma^2)$ per bounding box coordinate, replacing YOLO's deterministic box regression. 

Both heads operate per detection anchor, while fusion requires a single opinion per stream per frame. This frame-level opinion $(b_{UAV}, b_{bg}, u)$ and the spatial uncertainty are both read at the selected detection: the maximum-confidence NMS survivor, where confidence is the product of YOLO's objectness score and the UAV belief mass $b_{UAV}$. Rather than averaged over all anchors, since the overwhelming majority of anchors cover background and averaging would dilute the signal of the actual detection. The selected detection's per-coordinate variances are normalized by its predicted box dimensions for scale invariance and averaged into a single scalar $\hat{\sigma}^2_m$ per modality $m$. When a stream yields no detection, it contributes the vacuous opinion $b = (0, 0)$, $u = 1$, corresponding to the zero-evidence Dirichlet ($\alpha = \mathbf{1}$): a miss is treated as complete ignorance rather than as evidence for either class, so under belief fusion a missing stream cannot veto the other stream's prediction, it only increases the fused uncertainty.

\subsubsection{Uncertainty-aware classification fusion}

Because both sensors observe the same scene, the streams constitute dependent sources, making independence-assuming operators such as Dempster's rule \cite{dempster} inappropriate. While the Averaging Belief Fusion (ABF) rule \cite{Josang2017SL} handles dependent sources, it underestimates uncertainty when streams are confident but contradictory \cite{bezirganyan2025multimodallearninguncertaintyquantification}. This is a dangerous failure mode for safety-critical anti-UAV perception. This framework adopts Discounted Belief Fusion (DBF) \cite{bezirganyan2025multimodallearninguncertaintyquantification}, which introduces a conflict-based discounting step before ABF.

\textbf{Stage 1 — Discounting.} The inter-stream conflict $C = \sum_{j \neq k} b^{rgb}_j \cdot b^{th}_k$ yields a discounting factor $\eta = (1 - C^{\lambda})^{1/\lambda}$, where $\lambda$ controls penalty sharpness. Each stream's beliefs are rescaled: $\tilde{b}_m(c) = \eta \cdot b_m(c)$, with residual mass absorbed into uncertainty: $\tilde{u}_m = 1 - \sum_c \tilde{b}_m(c)$.

\textbf{Stage 2 — Fusion.} The discounted opinions are aggregated via ABF, assuming a uniform base rate $a_k = 1/K$:
$$b_{fused}(c) = \frac{\tilde{b}_{rgb}(c) \cdot \tilde{u}_{th} + \tilde{b}_{th}(c) \cdot \tilde{u}_{rgb}}{\tilde{u}_{rgb} + \tilde{u}_{th}}, \qquad u_{fused} = \frac{2 \cdot \tilde{u}_{rgb} \cdot \tilde{u}_{th}}{\tilde{u}_{rgb} + \tilde{u}_{th}}$$

The final class is $\hat{y} = \arg\max_c\; b_{fused}(c)$. When streams strongly conflict, $\eta \to 0$ drives both $\tilde{u}_m \to 1$, so $u_{fused} \to 1$: the system defaults to maximal uncertainty rather than a confidently wrong prediction. The fused uncertainty $u_{fused}$ serves as the uncertainty-gated output decision (Section~\ref{sec:edtc_switching}).

\subsubsection{Spatial fusion and uncertainty-gated output}
\label{sec:edtc_switching}

Due to misalignment between RGB and thermal frames in the Anti-UAV dataset, direct spatial fusion of bounding box coordinates is ill-posed: the modalities differ in resolution ($1920\!\times\!1080$ vs.\ $640\!\times\!512$) and lack spatial registration, so coordinates cannot be meaningfully combined across modalities. The framework therefore employs dynamic modality selection: at each frame, the bounding box from the modality with lower $\hat{\sigma}^2_m$ is selected as the system output \cite{he2019boundingboxregressionuncertainty}.


The uncertainty-gated commit decision of EDTC's switching mechanism is retained as an output gate: at each frame, the system emits the DMS-selected bounding box if and only if
$$b_{fused}(\text{UAV}) > b_{fused}(\text{bg}) \quad \text{AND} \quad u_{fused} < \theta_{cls},$$
and otherwise reports the target as absent. The class condition ensures the system only commits when the fused prediction favors the UAV class, since low semantic uncertainty alone could reflect a confident background prediction. The uncertainty condition guards against committing to near-vacuous predictions: the class condition alone is an argmax over beliefs and is satisfied even when both beliefs are negligible and almost all mass sits in $u_{fused}$. Requiring $u_{fused} < \theta_{cls}$ ensures the system only emits a box when the fused opinion carries sufficient evidence.

\subsection{Experimental Setup}

\textbf{Preprocessing.}
RGB frames are resized to $640{\times}640$ and
thermal frames are processed at native $640{\times}512$ resolution. Both modalities are  normalized with ImageNet statistics (mean $[0.485, 0.456, 0.406]$, std $[0.229, 0.224, 0.225]$). Detection augmentation follows standard YOLOv5s defaults (HSV jitter, translation, scaling). 

\textbf{Detection training.}
Both modality-specific YOLOv5s detectors are first pre-trained on their respective Anti-UAV training splits (IR and RGB) using standard YOLOv5s training for 20 epochs, producing the baseline checkpoints used for all baselines. The evidential framework is then added in two stages. In \emph{Stage~1} (Gaussian localization heads), Gaussian variance heads are appended to the frozen baseline and trained using the Gaussian negative log-likelihood loss:$$\mathcal{L}_{bbox} = \sum_{i \in \{x,y,w,h\}} \left[ \frac{(t_i - \mu_i)^2}{2\sigma_i^2} + \frac{1}{2}\log \sigma_i^2 \right]$$
where $t_i$ is the regression target and $\mu_i$, $\sigma_i^2$ are the predicted mean and variance for coordinate $i$. The first term penalizes prediction error scaled by predicted variance; the second prevents trivially predicting infinite variance. The Gaussian heads are warmed up for 5 epochs before full fine-tuning for 25 epochs (batch size 64, AdamW, weight decay $10^{-4}$, lr $10^{-3}$ then $10^{-4}$, cosine-annealed to $1\%$, Gaussian NLL weight $\lambda_\text{var}{=}0.1$). In \emph{Stage~2} (evidential classification heads), the Stage~1 network is frozen and only the Dirichlet evidence head is trained for 20 epochs (batch size 64, lr $10^{-3}$, cosine-annealed, $\lambda_\text{evi}{=}1.0$), with KL regularization annealed linearly from 0 to 1 over the first 10 epochs. Standard YOLO detection loss weights are used throughout: $\lambda_\text{box}{=}0.05$, $\lambda_\text{obj}{=}1.0$, $\lambda_\text{cls}{=}0.5$. All stages are run with three random seeds $\{0, 1, 2\}$. Results are reported as mean~$\pm$~std throughout.

\textbf{Detection inference.}
At inference, detections are formed with standard YOLOv5s post-processing: confidence threshold $0.25$, NMS IoU threshold $0.45$, and at most 300 detections per frame, with NMS restricted to the UAV class. The maximum-confidence survivor constitutes the selected detection from which the frame-level opinion and spatial uncertainty are read (Section~\ref{sec:architecture}). If no detection survives, the stream
contributes the vacuous opinion.

\textbf{Baselines.}
Five baselines are evaluated. \textit{Thermal-only detector} and \textit{RGB-only detector} run the single-stream architecture on each modality independently. \textit{Naive Late Fusion} averages the per-stream belief masses, $b_{fused} = \tfrac{1}{2}(b_{rgb} + b_{ir})$, without any uncertainty-aware reweighting. \textit{Gated Fusion} replaces the fixed average with a learned scalar gate $g \in (0,1)$, produced by a small MLP applied to the concatenated belief vectors of both streams: $b_{fused} = g \cdot b_{rgb} + (1-g) \cdot b_{ir}$. This allows the model to learn which modality to trust from data, but without explicit conflict modeling~\cite{bezirganyan2025multimodallearninguncertaintyquantification}. The gate MLP (${\approx}100$ parameters) is trained for 20 epochs with all backbone weights frozen, minimizing binary cross-entropy against frame-level UAV presence labels. \textit{Evidential Fusion} (ABF) applies the dual-stream evidential pipeline with $\eta{=}1$ (no conflict discounting), isolating the contribution of the DBF discounting mechanism. All fusion variants thus consume identical per-stream evidential outputs and differ only in the fusion operator.

\textbf{Threshold calibration.}
$\theta_{cls}$ is selected by an offline sweep over $\{0.05, 0.10, \ldots, 0.95\}$ on the validation set. The threshold maximizing Acc is applied unchanged on the test set. Thresholds are calibrated independently per method. For the naive and gated baselines, which produce no explicit uncertainty mass, the gating signal is defined as $1 - b_{fused}(\text{UAV})$. Additionally, the DBF discount exponent $\lambda$ is swept over {0.25, 0.5, 1, 2, 4} on the validation set. $\lambda=1$ is retained as the default.


\subsection{Evaluation Approach}


\textbf{Detection performance.}
We report five metrics. Intersection over Union (IoU): the ratio of overlapping to the union area of predicted and ground-truth boxes. This underlies all localization scores. AP@0.5 (Average Precision at IoU threshold 0.5) and mAP@[0.5:0.95] (AP averaged over thresholds 0.5–0.95 in steps of 0.05) are the primary detection metrics. Mean IoU over present frames (mIoU) isolates box-level quality independently of presence decisions. The EDTC Acc metric \cite{Zhu2023EDTC} ($\alpha{=}0.2$, $\beta{=}0.3$) combines per-frame IoU with a missed-detection penalty, included for comparability with EDTC/EDTC$^*$ baselines. The sparsification error AUSE$_\sigma$ ranks frames by spatial uncertainty $\hat{\sigma}^2_{best}$ and measures the area under the gap to an IoU-ordered oracle \cite{ilg2018uncertaintyestimatesmultihypothesesnetworks, geifman2017selectiveclassificationdeepneural, he2019boundingboxregressionuncertainty}. Full sparsification curves are reported alongside the scalar. Baselines:  thermal-only, RGB-only, Evidential Fusion (ABF), Naive Late Fusion and Gated Fusion.

\textbf{Uncertainty quality.}
We assess two distinct properties of the fused uncertainty. \emph{Calibration}: whether confidence values match empirical accuracy, via ECE (10 bins) and reliability diagrams \cite{guo2017calibrationmodernneuralnetworks}. \emph{Discrimination}: whether uncertainty separates correct from incorrect predictions, via AUROC for failure detection \cite{hendrycks2018baselinedetectingmisclassifiedoutofdistribution} on $u_{fused}$. Furthermore, for an uncertainty separability analysis, we evaluate both $u_{fused}$ and $\hat{\sigma}^2_{best}$ as failure predictors via AUROC.

\textbf{Presence discrimination.}
As the dataset is presence-dominated (imbalance $\approx44{:}1$ val, $73{:}1$ test), we assess how the fused belief separates present from absent frames using precision, recall, specificity, and balanced accuracy, computed from the belief alone (IoU-free). Specificity and balanced accuracy are emphasized as they expose the minority-class behavior that raw accuracy hides.

\textbf{Attribute-based robustness} Each sequence carries one or more of the seven attributes defined in Section \ref{sec:method-data}, plus TC difficulty tiers; subsets therefore overlap. For each attribute we pool the frames of all sequences carrying it and report detection performance (Acc), calibration (ECE) and uncertainty quality (Spearman rank correlation between per-frame $u_{fused}$ and IoU \cite{he2019boundingboxregressionuncertainty}, mean $u_{fused}$). 

\textbf{Computational efficiency} is measured via FPS, parameter count, and FLOPs, targeting $\geq$25 FPS on an NVIDIA A100 GPU (Snellius) \cite{reddi2020mlperfinferencebenchmark}.

\section{Results}
\label{sec:results}


\subsection{Detection Performance}
\label{sec:localization}

The analysis evaluates whether conflict-aware DBF improves detection performance compared with single-modality and undiscounted fusion baselines. Table \ref{tab:loc} addresses this via AP@0.5, mAP, mIoU, Acc and $AUSE_\sigma$. All fused variants exceed both single-modality streams on AP@0.5, mIoU, and Acc on both splits, and surpass the published thermal-only EDTC baselines on Acc (0.617/0.634) on the same Anti-UAV benchmark. On test, the fused operators reach Acc $0.670$ and mIoU $0.717$, against $0.604$/$0.664$ (IR) and $0.598$/$0.660$ (RGB). Within the fused group, ABF and DBF are identical on every metric; Naive matches them on all metrics except mAP, where it is lower by $0.001$; Gated coincides on mIoU and Acc; and all four belief-fusion variants tie on $\text{AUSE}_\sigma$ ($0.0497$ test). 

Table~\ref{tab:conflict} shows that inter-modality conflict $C$ never exceeds $1.2\times10^{-7}$ on any frame, leaving DBF's discounting step inert. A separate effect makes ABF indistinguishable from Naive averaging: on both-detect frames the maximum $|b_{\text{ABF}} - b_{\text{Naive}}|$ is $3\times10^{-3}$ (mean $1.6\times10^{-4}$), with zero class-label disagreements across 60k frames, because both streams are near-saturated (mean $b_{UAV}\approx0.99$).

\begin{figure}[htbp]
    \centering
    \includegraphics[width=1\linewidth]{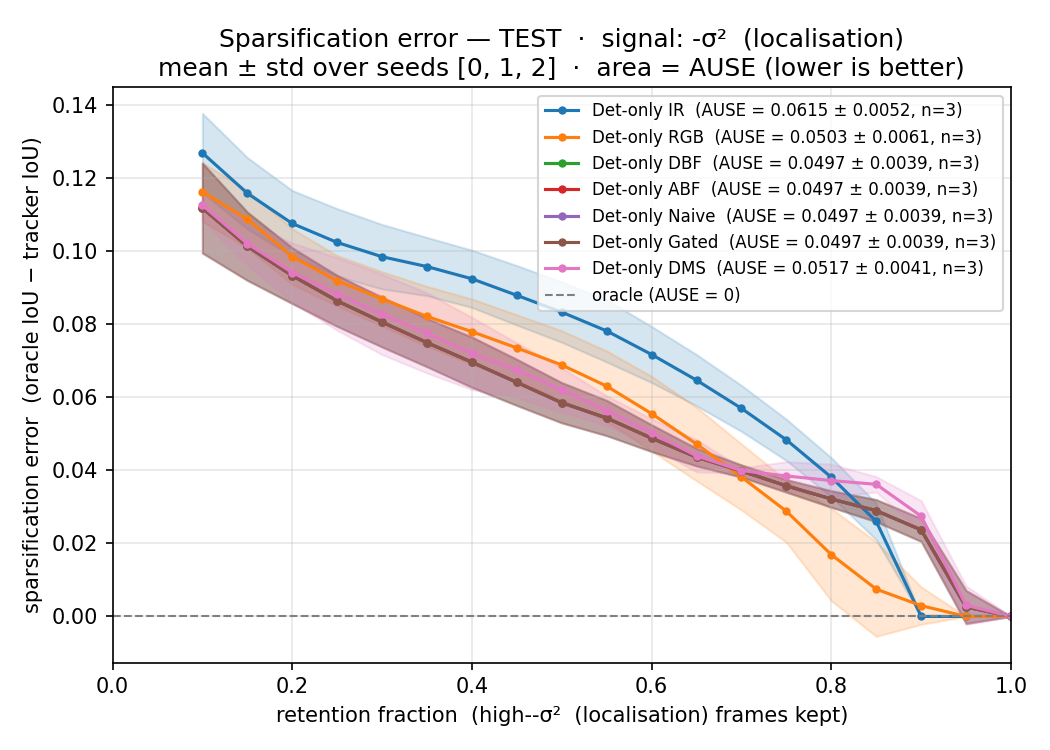}
    \caption{Sparsification curves on the test set (signal: $-\hat{\sigma}^2$, mean $\pm$ std over seeds 0-2). Each curve shows the gap between an IoU-ordered oracle and the model when frames are removed in order of \emph{decreasing} spatial uncertainty. Area under this gap is $\text{AUSE}_\sigma$ (lower is better). IR's Gaussian head is a substantially weaker localization failure predictor across the retention range (AUSE 0.0615), reflecting its weaker spatial uncertainty signal (Table \ref{tab:auroc-separability}). All four fusion operators are indistinguishable (AUSE = 0.0497), confirming the near-zero inter-modal conflict reported in Table \ref{tab:conflict}. DMS is marginally worse (0.0517).}
    \label{fig:spars}
\end{figure}

DMS posts the highest mAP ($0.480$ test) but ranks detections by $-\hat{\sigma}^2_{best}$ rather than Dirichlet expected probability, making its AP and mAP values not rank-comparable to the belief-based rows (see footnote). On mIoU and Acc it matches the fused variants; on $\text{AUSE}_\sigma$ it is marginally worse ($0.0517$ vs.\ $0.0497$). Figure \ref{fig:spars} shows the full sparsification curves on the test set: IR performs substantially worse than all other methods across the retention range, while DBF, ABF, Naive and Gated overlap exactly throughout. 

Additionally, a sweep of the DBF discount exponent over $\lambda \in \{0.25, 0.5, 1, 2, 4\}$ (validation, seeds 0--2) leaves every metric unchanged across all values.

\begin{table*}[htbp]
\centering
\caption{Localization performance and benchmark comparability (mean $\pm$ std over seeds 0--2).
AP@0.5 and mAP@[0.5:0.95] are the primary detection metrics; mIoU is mean IoU over present frames; Acc is the EDTC accuracy metric included for comparability with the original thermal-only baselines (det+track); AUSE$_\sigma$ is the sparsification error using $-\hat{\sigma}^2_{best}$ (lower is better). \textbf{Bold} = best within each group per split.}
\label{tab:loc}
\begin{tabular}{clllccccc}
\toprule
Split & Type & Model
  & AP@0.5 $\uparrow$
  & mAP $\uparrow$
  & mIoU $\uparrow$
  & EDCT's Acc $\uparrow$
  & AUSE$_\sigma$ $\downarrow$ \\
\midrule
\multirow{7}{*}{\rotatebox[origin=c]{90}{\textit{Val.}}}
 & \multirow{7}{*}{\shortstack[l]{Proposed\\(Det-only)}}
 & IR
   & $0.944 \pm 0.022$ & $0.531 \pm 0.014$
   & $0.765 \pm 0.017$ & $0.732 \pm 0.027$
   & $0.0593 \pm 0.0005$ \\
 && RGB
   & $0.907 \pm 0.043$ & $0.596 \pm 0.034$
   & $0.782 \pm 0.033$ & $0.730 \pm 0.057$
   & $0.0443 \pm 0.0011$ \\
 && ABF
   & $\mathbf{0.960 \pm 0.011}$ & $\mathbf{0.598 \pm 0.015}$
   & $\mathbf{0.806 \pm 0.009}$ & $\mathbf{0.779 \pm 0.020}$
   & $\mathbf{0.0434 \pm 0.0028}$ \\
 && DBF
   & $\mathbf{0.960 \pm 0.011}$ & $\mathbf{0.598 \pm 0.015}$
   & $\mathbf{0.806 \pm 0.009}$ & $\mathbf{0.779 \pm 0.020}$
   & $\mathbf{0.0434 \pm 0.0028}$ \\
 && Naive
   & $\mathbf{0.960 \pm 0.011}$ & $0.597 \pm 0.016$
   & $\mathbf{0.806 \pm 0.009}$ & $\mathbf{0.779 \pm 0.020}$
   & $\mathbf{0.0434 \pm 0.0028}$ \\
 && Gated
   & $0.958 \pm 0.008$ & $0.594 \pm 0.010$
   & $\mathbf{0.806 \pm 0.009}$ & $\mathbf{0.779 \pm 0.020}$
   & $\mathbf{0.0434 \pm 0.0029}$ \\
 && DMS
   & $0.959 \pm 0.013^{\dagger}$ & $0.604 \pm 0.015^{\dagger}$
   & $0.804 \pm 0.012$ & $0.777 \pm 0.019$
   & $0.0456 \pm 0.0030$ \\
\midrule
\multirow{9}{*}{\rotatebox[origin=c]{90}{\textit{Test}}}
 & \multirow{2}{*}{\shortstack[l]{Ref.\\(Det+Track)}}
 & EDTC~\cite{Zhu2023EDTC}      & --- & --- & --- & 0.617 & --- \\
 && EDTC$^*$~\cite{Zhu2023EDTC}  & --- & --- & --- & 0.634 & --- \\
\cmidrule{2-8}
 & \multirow{7}{*}{\shortstack[l]{Proposed\\(Det-only)}}
 & IR
   & $0.832 \pm 0.009$ & $0.409 \pm 0.003$
   & $0.664 \pm 0.005$ & $0.604 \pm 0.008$
   & $0.0615 \pm 0.0043$ \\
 && RGB
   & $0.811 \pm 0.044$ & $0.442 \pm 0.029$
   & $0.660 \pm 0.033$ & $0.598 \pm 0.044$
   & $0.0503 \pm 0.0049$ \\
 && ABF
   & $\mathbf{0.888 \pm 0.007}$ & $\mathbf{0.470 \pm 0.010}$
   & $\mathbf{0.717 \pm 0.006}$ & $\mathbf{0.670 \pm 0.010}$
   & $\mathbf{0.0497 \pm 0.0039}$ \\
 && DBF
   & $\mathbf{0.888 \pm 0.007}$ & $\mathbf{0.470 \pm 0.010}$
   & $\mathbf{0.717 \pm 0.006}$ & $\mathbf{0.670 \pm 0.010}$
   & $\mathbf{0.0497 \pm 0.0039}$ \\
 && Naive
   & $\mathbf{0.888 \pm 0.007}$ & $0.469 \pm 0.010$
   & $\mathbf{0.717 \pm 0.006}$ & $\mathbf{0.670 \pm 0.010}$
   & $\mathbf{0.0497 \pm 0.0039}$ \\
 && Gated
   & $0.884 \pm 0.006$ & $0.466 \pm 0.008$
   & $\mathbf{0.717 \pm 0.006}$ & $\mathbf{0.670 \pm 0.010}$
   & $\mathbf{0.0497 \pm 0.0038}$ \\
 && DMS
   & $0.885 \pm 0.007^{\dagger}$ & $0.480 \pm 0.008^{\dagger}$
   & $\mathbf{0.717 \pm 0.007}$ & $0.669 \pm 0.010$
   & $0.0517 \pm 0.0034$ \\
\bottomrule
\end{tabular}
\smallskip

\raggedright
\small $^{\dagger}$DMS produces no fused belief; its detections are ranked by
localization variance ($-\hat{\sigma}^2_{best}$) rather than the Dirichlet
expected probability used for all other rows. AP values are therefore not
rank-comparable to the belief-based variants and are reported for reference only.
\end{table*}

\begin{table}[htbp]
\centering
\caption{Frame taxonomy and inter-modality conflict $C=\sum_{j\neq k} b^{\mathrm{rgb}}_j\, b^{\mathrm{ir}}_k$, computed from the per-stream beliefs entering fusion (test split; values are the maximum over frames in each row, with ranges spanning seeds 0--2).}
\label{tab:conflict}
\small
\begin{tabular}{lcc}
\toprule
Situation & Share of frames & Conflict $C$ \\
\midrule
Both streams detect & 71.5--79.1\% & $\leq 1.2\times10^{-7}$ \\
One stream detects  & 16.2--21.5\% & $0$ \\
Neither detects     & 4.7--7.2\%   & ---  \\
\bottomrule
\end{tabular}
\end{table}

\subsection{Uncertainty Quality}
\label{sec:uncertainty}

The analysis further assesses how well-calibrated the fused uncertainty estimates are and whether they reliably discriminate between correct and incorrect predictions. The three subsections below address calibration, the operating range of $u_{fused}$, and its discriminative power in turn.

\subsubsection{Presence Discrimination \& Calibration}
\label{sec:uncertainty-calibration}
Table~\ref{tab:belief} evaluates the fused semantic belief independently of the box output. DMS produces no fused belief and is not assessed here.

Precision is uniformly high ($\geq$0.99) across all models. The unimodal streams achieve higher specificity: IR reaches $0.950$ on validation and RGB $0.944$ on test, whereas all four fusion operators cluster near $0.782$ val and $0.693$ test. Consequently, on balanced accuracy the unimodal streams lead each split (IR $0.961$ val, RGB $0.897$ test), with the fusion group trailing at $\approx$$0.887$ val and $0.821$ test. ABF, DBF, and Naive share identical presence-discrimination columns, differing only in the confidence score attached to each decision.

\begin{figure}[htbp]
    \centering
    \includegraphics[width=1\linewidth]{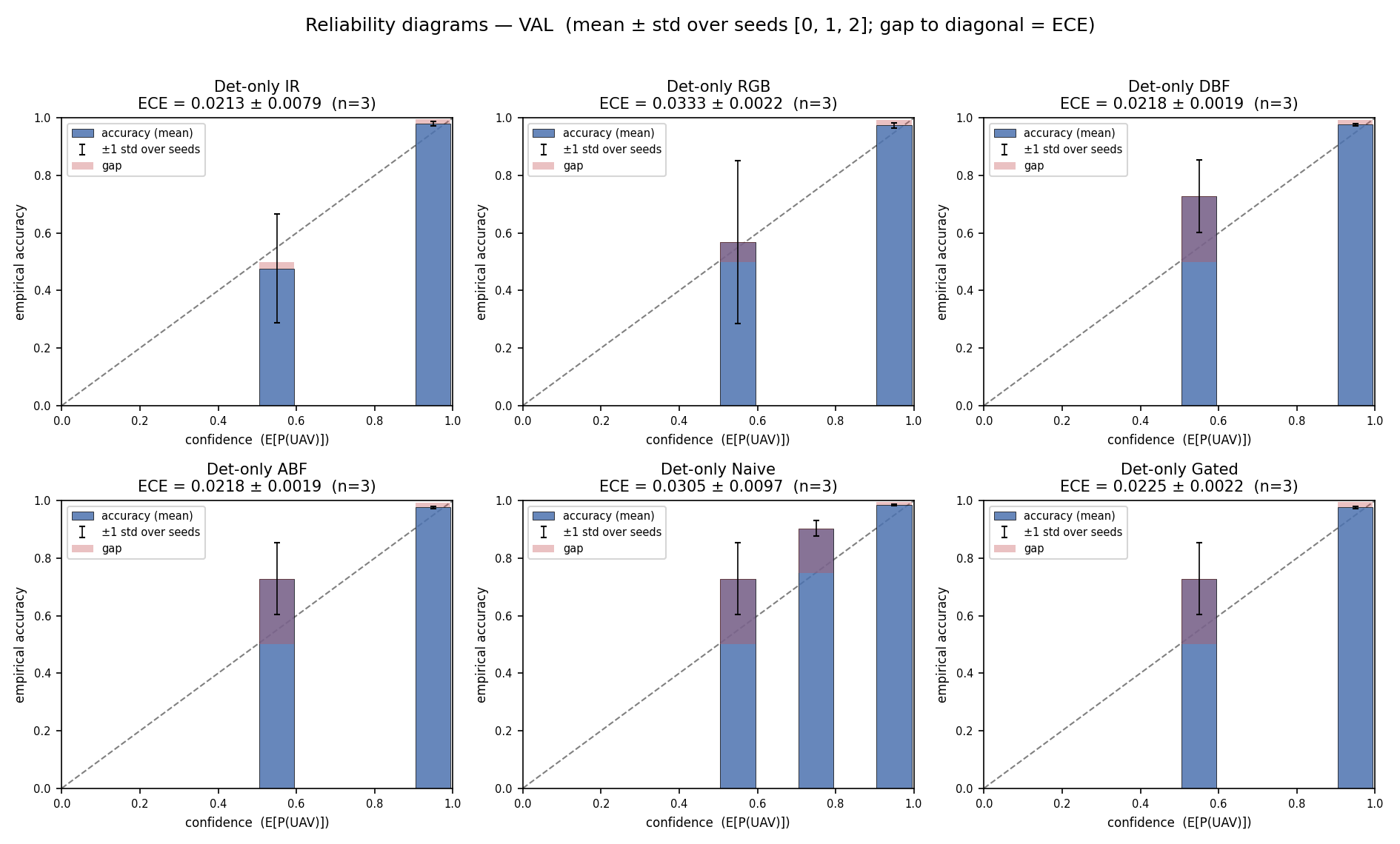}
    \caption{Reliability diagrams on the validation set (mean $\pm$ std over seeds 0-2). The gap between each bar and the diagonal equals the per-bin calibration error; ECE is the weighted mean gap. All methods concentrate mass in two bins: a mid-confidence cluster $(\approx0.55)$ from harder frames where the detector confidence is lower, and a high-confidence cluster $(\approx0.95-1.0)$ from the dominant presence frames. At the mid-confidence bin, IR is overconfident (bar below diagonal), while fusion operators are underconfident (bar above diagonal); RGB is best calibrated there. Naive exhibits an additional bin near 0.75 from missing-detection frames encoded with $u \approx 0.5$ (Table \ref{tab:theta-sweep}). DBF and ABF are identical.}
    \label{fig:reliab}
\end{figure}

On calibration (ECE), fusion operators are competitive on test: Naive $0.055$, ABF/DBF $0.057$, Gated $0.058$; IR is the worst-calibrated model ($0.082$). On classification failure discrimination (AUROC, whether the fused belief separates correctly and incorrectly classified presence/absence frames), unimodal streams lead on test (IR $0.908$, RGB $0.912$), with fusion operators close behind (ABF/DBF/Naive $0.892$, Gated $0.858$). ABF and DBF are indistinguishable on every metric. Figure \ref{fig:reliab} shows the reliability diagrams on the validation set, all of the methods show two populated confidence bins near 0.55 and 1.0, with DBF and ABF producing identical diagrams. Naive exhibits an additional bin near 0.75.

\begin{table*}[htbp]
\centering
\caption{Fused-belief quality, evaluated independently of the box output.
Presence-discrimination metrics use the fused semantic belief only (IoU-free) under
heavy class imbalance (val $\approx$44:1, test $\approx$73:1 present:absent).
ECE measures calibration of the belief values; AUROC measures failure
(misclassification) discrimination. DMS produces no fused belief and is not assessed
here. Results are mean\,$\pm$\,std over seeds 0-2. \textbf{Bold} = best per column within each split.}
\label{tab:belief}
\setlength{\tabcolsep}{3pt}
\begin{tabular}{clcccccc}
\toprule
 & & \multicolumn{4}{c}{Presence discrimination} & \multicolumn{2}{c}{Uncertainty quality} \\
\cmidrule(lr){3-6}\cmidrule(lr){7-8}
Split & Model
  & Prec $\uparrow$
  & Recall $\uparrow$
  & Spec $\uparrow$
  & BalAcc $\uparrow$
  & ECE $\downarrow$
  & AUROC $\uparrow$ \\
\midrule
\multirow{6}{*}{\rotatebox[origin=c]{90}{\textit{Validation}}}
 & IR
  & \textbf{0.9988\,$\pm$\,0.0004}
  & 0.9719\,$\pm$\,0.0163
  & \textbf{0.9497\,$\pm$\,0.0195}
  & \textbf{0.9608\,$\pm$\,0.0059}
  & \textbf{0.0213\,$\pm$\,0.0078}
  & \textbf{0.8770\,$\pm$\,0.0562} \\
 & RGB
  & 0.9900\,$\pm$\,0.0005
  & 0.9451\,$\pm$\,0.0423
  & 0.8494\,$\pm$\,0.0084
  & 0.8973\,$\pm$\,0.0197
  & 0.0333\,$\pm$\,0.0022
  & 0.7777\,$\pm$\,0.1320 \\
 & ABF
  & 0.9950\,$\pm$\,0.0005
  & \textbf{0.9929\,$\pm$\,0.0041}
  & 0.7820\,$\pm$\,0.0192
  & 0.8874\,$\pm$\,0.0085
  & 0.0218\,$\pm$\,0.0019
  & 0.7557\,$\pm$\,0.0439 \\
 & DBF
  & 0.9950\,$\pm$\,0.0004
  & \textbf{0.9929\,$\pm$\,0.0041}
  & 0.7820\,$\pm$\,0.0192
  & 0.8874\,$\pm$\,0.0085
  & 0.0218\,$\pm$\,0.0019
  & 0.7558\,$\pm$\,0.0439 \\
 & Naive
  & 0.9950\,$\pm$\,0.0004
  & \textbf{0.9929\,$\pm$\,0.0041}
  & 0.7820\,$\pm$\,0.0192
  & 0.8874\,$\pm$\,0.0085
  & 0.0305\,$\pm$\,0.0097
  & 0.7621\,$\pm$\,0.0468 \\
 & Gated
  & 0.9950\,$\pm$\,0.0004
  & 0.9921\,$\pm$\,0.0042
  & 0.7817\,$\pm$\,0.0160
  & 0.8873\,$\pm$\,0.0070
  & 0.0225\,$\pm$\,0.0018
  & 0.7156\,$\pm$\,0.0767 \\
\midrule
\multirow{6}{*}{\rotatebox[origin=c]{90}{\textit{Test}}}
 & IR
  & \textbf{0.9983\,$\pm$\,0.0001}
  & 0.8945\,$\pm$\,0.0081
  & 0.8907\,$\pm$\,0.0092
  & 0.8926\,$\pm$\,0.0054
  & 0.0815\,$\pm$\,0.0055
  & 0.9084\,$\pm$\,0.0028 \\
 & RGB
  & 0.9954\,$\pm$\,0.0009
  & 0.8509\,$\pm$\,0.0550
  & \textbf{0.9439}\,$\pm$\,0.0141
  & \textbf{0.8974}\,$\pm$\,0.0205
  & 0.0574\,$\pm$\,0.0202
  & \textbf{0.9115\,$\pm$\,0.0081} \\
 & ABF
  & 0.9956\,$\pm$\,0.0003
  & \textbf{0.9481\,$\pm$\,0.0126}
  & 0.6930\,$\pm$\,0.0221
  & 0.8206\,$\pm$\,0.0085
  & 0.0572\,$\pm$\,0.0006
  & 0.8916\,$\pm$\,0.0069 \\
 & DBF
  & 0.9956\,$\pm$\,0.0003
  & \textbf{0.9481\,$\pm$\,0.0126}
  & 0.6930\,$\pm$\,0.0221
  & 0.8206\,$\pm$\,0.0085
  & 0.0572\,$\pm$\,0.0006
  & 0.8916\,$\pm$\,0.0069 \\
 & Naive
  & 0.9956\,$\pm$\,0.0003
  & \textbf{0.9481\,$\pm$\,0.0126}
  & 0.6930\,$\pm$\,0.0221
  & 0.8206\,$\pm$\,0.0085
  & \textbf{0.0547\,$\pm$\,0.0150}
  & 0.8919\,$\pm$\,0.0060 \\
 & Gated
  & 0.9956\,$\pm$\,0.0003
  & \textbf{0.9481\,$\pm$\,0.0101}
  & 0.6930\,$\pm$\,0.0018
  & 0.8206\,$\pm$\,0.0070
  & 0.0584\,$\pm$\,0.0007
  & 0.8577\,$\pm$\,0.0113 \\
\bottomrule
\end{tabular}
\end{table*}

\subsubsection{Calibration of Threshold $\theta_{\mathrm{cls}}$}
\label{sec:results-theta-sweep}
Table \ref{tab:theta-sweep} reports the $\theta_{\mathrm{cls}}$ sweep results. In no configuration does applying $\theta_{\mathrm{cls}}$ improve over the thresholdless $\arg\max_c b_{\mathrm{fused}}(c)$ decision rule. The resulting Acc values are those reported in Table \ref{tab:loc}.

For DBF and ABF, $u_{\mathrm{fused}}$ on gate-eligible frames peaks at $0.046$; for Gated it peaks at $0.023$. Both are below the lowest grid value $0.05$. the gate therefore never fires and reduces exactly to $\arg\max_c b_{\mathrm{fused}}(c)$ across all three variants and all three seeds.

Naive is the sole exception: missing-detection frames enter fusion with $u_{\mathrm{fused}} \approx 0.5$, making up 4.8--13.5\% of gate-eligible frames. The gate fires, but yields no gain: accuracy plateaus at the $\arg\max$ value for $\theta_{\mathrm{cls}} \geq 0.55$ and degrades below it.

\begin{table}[htbp]
\centering
\caption{$\theta_{\mathrm{cls}}$ sweep (validation). $u_{\mathrm{fused}}$
statistics on gate-eligible frames ($n \approx 60$k); ranges span seeds $0$--$2$.
For DBF, ABF and Gated the threshold is inert (gate $\equiv \arg\max b_{\mathrm{fused}}$)
and Acc is constant over the entire grid. For Naive the threshold is active but
yields no gain: Acc plateaus at the $\arg\max$ value for $\theta_{\mathrm{cls}}\ge0.55$
and degrades below it.}
\label{tab:theta-sweep}
\small
\begin{tabular}{lccc}
\toprule
Model & median $u_{\mathrm{fused}}$ & max $u_{\mathrm{fused}}$ & $\theta_{\mathrm{cls}}$ effect \\
\midrule
DBF   & 0.009--0.011 & 0.046 & inert \\
ABF   & 0.009--0.011 & 0.046 & inert \\
Gated & 0.009--0.010 & 0.023 & inert \\
Naive & 0.009--0.011 & 0.51  & active ($\theta^\star\!\ge\!0.55$) \\
\bottomrule
\end{tabular}
\end{table}

\subsubsection{Semantic vs. Spatial Uncertainty Comparison}
\label{sec:separability}

Whereas the AUROC in Table \ref{tab:belief} measures classification failure (presence/absence), here the question shifts to localization failure: among present frames where the fused prediction already favors UAV, does $u_{\text{fused}}$ or $\hat{\sigma}^2_{\text{best}}$ better predict whether localization succeeded? Table~\ref{tab:auroc-separability} reports failure-detection AUROC for $u_{\text{fused}}$ and $\hat{\sigma}^2_{\text{best}}$ on present frames of the validation set. In RGB and every multimodal configuration, $\hat{\sigma}^2_{\text{best}}$ is the stronger failure indicator (0.748 RGB, $0.739$ for fusion variants), while $u_{\text{fused}}$ ranges from $0.626$ (DBF/ABF) to $0.676$ (Gated). The IR-only stream is the sole exception: there $\hat{\sigma}^2_{\text{best}}$ ($0.560$) is near-chance and $u_{\text{fused}}$ ($0.757$) is the stronger signal, reversing the ordering seen in every other configuration.

\begin{table}[htbp]
\centering
\caption{Failure-detection AUROC on the validation set, evaluated per signal across all six models. Evaluation is restricted to present frames conditioned on $\arg\max = \text{UAV}$; a failure is defined as $\text{IoU} < 0.5$. AUROC $> 0.5$ indicates the signal ranks failure frames above success frames; AUROC $= 0.5$ is random; AUROC $< 0.5$ indicates the signal ranks failures \emph{below} successes. \textbf{Bold} denotes the higher AUROC per configuration.}
\label{tab:auroc-separability}
\begin{tabular}{lcc}
\toprule
& \multicolumn{2}{c}{Failure-detection AUROC $\uparrow$} \\
\cmidrule(lr){2-3}
Model & $u_{\text{fused}}$ & $\hat{\sigma}^2_{\text{best}}$ \\
\midrule
IR    & $\mathbf{0.757 \pm 0.083}$ & $0.560 \pm 0.015$ \\
RGB   & $0.657 \pm 0.066$ & $\mathbf{0.748 \pm 0.060}$ \\
\midrule
DBF   & $0.626 \pm 0.062$ & $\mathbf{0.739 \pm 0.045}$ \\
ABF   & $0.626 \pm 0.062$ & $\mathbf{0.739 \pm 0.045}$ \\
Naive & $0.635 \pm 0.070$ & $\mathbf{0.739 \pm 0.045}$ \\
Gated & $0.676 \pm 0.086$ & $\mathbf{0.739 \pm 0.045}$ \\
\bottomrule
\end{tabular}
\end{table}

\subsection{Attribute-based Robustness}
\label{sec:robustness}

The analysis further examines whether uncertainty-aware fusion degrades more gracefully than single-modality baselines under challenging conditions. Table~\ref{tab:attribute_test} reports Acc, ECE, and uncertainty--failure coupling for each attribute subset.

Fusion achieves the highest Acc on every attribute subset. The thermal-crossover tiers show a monotone decline ($0.773 \rightarrow 0.667 \rightarrow 0.463$ for fusion), with TC-Hard as the hardest condition across all methods.

Fusion attains the lowest ECE on 8 of 10 subsets. TC-Hard is the worst-calibrated condition for all methods (ECE $\geq 0.141$). The OC subset rests on $n{=}2$ sequences and should be read as indicative only.

RGB yields the most negative $\rho(u, \text{IoU})$ on 9 of 10 subsets. Fusion is consistently weaker, with $\rho$ between $-0.036$ and $-0.579$. The diagnostic $\bar{u}$ shows fusion operates at a markedly lower uncertainty scale ($0.017$--$0.206$) than either unimodal stream. IR's $\rho$ turns positive under low illumination ($+0.283$).

\begin{table*}[htbp]
\centering
\caption{Attribute-based robustness on the Anti-UAV test set. Means over seeds 0-2. $\rho$ is Spearman correlation between per-frame $u_\text{fused}$ and IoU (more negative = uncertainty better tracks localization failure). $\bar{u}$ is mean $u_\text{fused}$ over all frames (including non-detections), reported as a diagnostic of each method's uncertainty scale (not a quality metric; not scored). Best per row within each scored metric group in \textbf{bold}.}
\label{tab:attribute_test}
\setlength{\tabcolsep}{4pt}
\begin{tabular}{l ccc c ccc c ccc c ccc}
\toprule
& \multicolumn{3}{c}{EDTC's Acc $\uparrow$} & & \multicolumn{3}{c}{ECE $\downarrow$} & & \multicolumn{3}{c}{$\rho(u,\text{IoU})$ $\downarrow$} & & \multicolumn{3}{c}{$\bar{u}$ (diag.)} \\
\cmidrule(lr){2-4} \cmidrule(lr){6-8} \cmidrule(lr){10-12} \cmidrule(lr){14-16}
Attr & IR & RGB & ABF/DBF & & IR & RGB & ABF/DBF & & IR & RGB & ABF/DBF & & IR & RGB & ABF/DBF \\
\midrule
ALL & 0.605 & 0.598 & \textbf{0.670} & & 0.081 & 0.057 & \textbf{0.057} & & $-0.339$ & $\mathbf{-0.442}$ & $-0.288$ & & 0.126 & 0.211 & 0.073 \\
TC  & 0.560 & 0.553 & \textbf{0.632} & & 0.110 & 0.083 & \textbf{0.077} & & $-0.425$ & $\mathbf{-0.524}$ & $-0.347$ & & 0.156 & 0.247 & 0.092 \\
FM  & 0.618 & 0.576 & \textbf{0.660} & & 0.078 & 0.070 & \textbf{0.066} & & $-0.135$ & $\mathbf{-0.460}$ & $-0.242$ & & 0.101 & 0.200 & 0.066 \\
SV  & 0.533 & 0.508 & \textbf{0.584} & & 0.108 & 0.086 & \textbf{0.094} & & $-0.371$ & $\mathbf{-0.593}$ & $-0.381$ & & 0.181 & 0.325 & 0.129 \\
LI  & 0.681 & 0.603 & \textbf{0.705} & & 0.045 & 0.062 & \textbf{0.045} & & $+0.283$ & $\mathbf{-0.400}$ & $-0.036$ & & 0.035 & 0.129 & 0.017 \\
LR  & 0.598 & 0.517 & \textbf{0.622} & & 0.078 & 0.077 & \textbf{0.080} & & $-0.019$ & $\mathbf{-0.528}$ & $-0.190$ & & 0.088 & 0.274 & 0.062 \\
OV  & 0.673 & 0.550 & \textbf{0.678} & & \textbf{0.028} & 0.048 & 0.045 & & $-0.162$ & $\mathbf{-0.456}$ & $-0.215$ & & 0.120 & 0.313 & 0.096 \\
OC*  & 0.631 & 0.572 & \textbf{0.702} & & \textbf{0.086} & 0.049 & 0.093 & & $-0.412$ & $\mathbf{-0.458}$ & $-0.162$ & & 0.196 & 0.384 & 0.138 \\
\midrule
\multicolumn{16}{l}{\textit{Thermal crossover by difficulty}} \\
TC-Easy & 0.657 & 0.688 & \textbf{0.773} & & 0.059 & 0.033 & \textbf{0.003} & & $\mathbf{-0.351}$ & $-0.277$ & $-0.077$ & & 0.106 & 0.082 & 0.017 \\
TC-Mid  & 0.612 & 0.576 & \textbf{0.667} & & 0.070 & 0.075 & \textbf{0.053} & & $-0.288$ & $\mathbf{-0.471}$ & $-0.255$ & & 0.102 & 0.215 & 0.060 \\
TC-Hard & 0.395 & 0.408 & \textbf{0.463} & & 0.218 & \textbf{0.141} & 0.178 & & $-0.630$ & $\mathbf{-0.748}$ & $-0.579$ & & 0.288 & 0.428 & 0.206 \\
\bottomrule
\end{tabular}
\smallskip

\raggedright
\small *OC (occlusion) contains only $n{=}2$ test sequences; values are high-variance and should be read as indicative, not conclusive.
\end{table*}

\subsection{Computational Efficiency}
\label{sec:efficiency}

The analysis further evaluates whether the dual-stream architecture meets real-time constraints and whether the choice of fusion operator affects throughput. Table~\ref{tab:efficiency} reports the parameter count, GFLOPs, and inference speed (FPS). All variants exceed the 25 FPS real-time threshold, with unimodal detectors achieving approximately 80 FPS and multimodal variants operating at approximately 38--41 FPS. The multimodal architectures approximately double the parameter count and GFLOPs relative to the unimodal streams, while inference speed remains comparable across the different fusion strategies.

\begin{table}[htbp]
\centering
\setlength{\tabcolsep}{4pt}   
\caption{Computational efficiency. Parameters
and GFLOPs are deterministic; FPS is reported as mean~$\pm$~std over seeds 0-2 on an NVIDIA A100 GPU (Snellius).}
\label{tab:efficiency}
\begin{tabular}{l r r c c}
\toprule
Model & Params (M) & GFLOPs & FPS (val) & FPS (test) \\
\midrule
IR  & 7.95  & 28.40 & $82.56\pm0.91$ & $81.05\pm4.07$ \\
RGB & 7.95  & 28.40 & $77.16\pm1.79$ & $77.75\pm1.44$ \\
Naive        & 15.89 & 56.80 & $38.39\pm0.98$ & $40.69\pm0.94$ \\
ABF          & 15.89 & 56.80 & $38.98\pm0.70$ & $39.30\pm0.58$ \\
DBF          & 15.89 & 56.80 & $39.67\pm0.07$ & $40.01\pm0.95$ \\
DMS          & 15.89 & 56.80 & $39.13\pm0.57$ & $39.99\pm1.21$ \\
Gated        & 15.89 & 56.80 & $38.83\pm1.40$ & $38.49\pm0.82$ \\
\bottomrule
\end{tabular}
\end{table}





\section{Discussion}
\label{sec:discussion}

\subsection{Detection Performance}
\label{sec:disc-pred}

The analysis examines whether conflict-aware DBF improves performance relative to single-stream and undiscounted fusion baselines. The results show that fusion consistently improves localization performance over either modality alone across the evaluated splits (Table~\ref{tab:loc}). All fused variants also outperform the published EDTC ($0.617$) and EDTC$^*$ ($0.634$) results.


However, the central finding is that no fusion operator outperforms the others. DBF is identical to ABF on every metric, and both match naive fusion on all decision metrics. This is structural: the near-zero conflict ($C \leq 1.2\times10^{-7}$, Table \ref{tab:conflict}) means the discounting factor collapses to $\eta \approx 1$ and DBF reduces exactly to ABF. A second, independent collapse explains why ABF in turn matches naive fusion. ABF equals the naive mean exactly when $(b_{RGB}-b_{th})(u_{RGB}-u_{th})=0$, thus when the streams carry equal belief or equal uncertainty. On the both-detect frames the streams are near-saturated (mean $b_{UAV}\approx0.99$ for both modalities), so $b_{RGB}-b_{th}\approx0$ and ABF's uncertainty weighting reweighs near-identical beliefs onto the same value. The maximum fused-belief gap to naive is $3 \times 10^{-3}$ and the two rules never disagree on the predicted class. Additionally, the sparsification curves (Figure \ref{fig:spars}) confirm this visually: all four fusion operators overlap exactly across the full retention range. IR remains substantially worse throughout, this is consistent with its outcome-insensitive Gaussian head discussed in Section \ref{sec:disc-uncertainty}. The near-zero conflict is not accidental, it follows from two independent properties of this benchmark and its modeling choices.


The first is a \textit{dataset property:} the Anti-UAV benchmark is presence-dominated, with the UAV visible in approximately $94-99\%$ of frames and absence segments lasting a median of 28 frames. On such a benchmark, when both streams detect, they tend to agree: both surface high $b_{UAV}$ anchors on the same target. This leaves negligible $b_{bg}$ mass for conflict to accumulate against. 

The second is a \textit{modeling choice:} non-detecting streams are encoded as the vacuous opinion ($b{=}(0,0)$, $u{=}1$). This is the correct Subjective Logic treatment of ignorance: a stream that fails to detect should not be interpreted as evidence for background. This prevents a temporarily blind modality from vetoing a true positive seen by the other stream, the more dangerous error in safety-critical perception. The consequence however, is that a non-detecting stream contributes $C = 0$ by construction, regardless of what the detecting stream believes. As Table \ref{tab:conflict} shows, 16-21\% of frames fall into the one-stream-detects category, and every such frame is excluded from generating conflict by the encoding itself. This mechanism is falsifiable on the existing dataset: routing low detector objectness into evidence for background would allow these frames to generate genuine conflict and activate the discounting step. Crucially, this does not reintroduce the veto behavior the vacuous encoding was designed to prevent (Section \ref{sec:architecture}). The two cases are distinct: a stream that produces no detection at all remains vacuous $(b = (0,0), u = 1)$ and still cannot veto the other stream, preserving the safety property under genuine blindness. Only a stream that does detect, but with low objectness, would contribute background belief. Conflict would then arise from weak-but-present disagreement rather than from absence. Objectness currently only multiplies $b_{UAV}$, so it can shrink UAV belief but never constitute evidence against a UAV. Aligning the encoding with this distinction is left for future work. 


The threshold sweep (Section~\ref{sec:results-theta-sweep}) shows the same pattern from a different angle. If conflict were meaningful, $u_{\mathrm{fused}}$ would vary enough to engage the gate. Instead, $u_{\mathrm{fused}}$ on gate-eligible frames peaks at 0.046: below the lowest sweep value. So the gate never fires for DBF, ABF and Gated. For Naive, the gate does fire, but only on missing-detection frames already encoded as vacuous, so it cannot improve over the arg max decision. The gate therefore adds nothing on this benchmark.

The $\lambda$ sweep (Section~\ref{sec:localization}) confirms this directly. Tuning the discount exponent over a wide range leaves every metric unchanged, because with $C \approx 0$ the discount factor $(1-C)^{\lambda} \approx 1$ for every $\lambda$, making the fused opinion independent of the exponent. We therefore retain the standard linear setting $\lambda=1.0$ as the principled default.

\subsection{Uncertainty Quality}
\label{sec:disc-uncertainty}




The analysis further assesses the calibration of the fused uncertainty estimates and their ability to reliably discriminate between correct and incorrect predictions. The fused semantic uncertainty is well-calibrated but is not the system's strongest failure detector. On calibration (ECE, Table \ref{tab:belief}), fusion operators are competitive on test (ECE 0.057 for ABF/DBF, 0.055 for Naive), outperforming IR (0.082) and matching RGB (0.057). DBF and ABF remain indistinguishable, consistent with the inert discounting step established in Section~\ref{sec:disc-pred}. On classification failure discrimination (AUROC, Table~\ref{tab:belief}), unimodal streams marginally lead on test (IR 0.908, RGB 0.912), with fusion operators close behind (ABF/DBF/Naive 0.892, Gated 0.858). The reliability diagrams (Figure \ref{fig:reliab}) make the calibration structure visible: at the mid-confidence bin $(\approx0.55)$, IR is overconfident while fusion operators are underconfident, but all methods are well calibrated at the high-confidence bin $(\approx 0.95-1.0)$.

The failure mode this benchmark stresses is itself diagnostic. The failure-detection evaluation on the comparison between semantic vs. spatial uncertainty (Table \ref{tab:auroc-separability}) is restricted to present frames on which the fused prediction already favors UAV, with failure defined as IoU < 0.5. That is, the dominant error is a detected target that is poorly localized, not a target misclassified as background. Because the errors are localization errors rather than classification errors, a spatial signal should naturally outperform a semantic one, which is what we observe: $\hat{\sigma}^2_{\text{best}}$ is the stronger failure signal for RGB and all multimodal configurations (AUROC 0.739-0.748 vs. 0.626-0.676 for $u_{\text{fused}}$). Calibration alone therefore overstates the usefulness of $u_{\text{fused}}$: $u_{\text{fused}}$ is additionally compressed into $[0.009, 0.046]$ on gate-eligible frames (Table \ref{tab:theta-sweep}) for fusion configurations, further limiting its discriminative range \cite{sackett2000range}.

However, the IR-only configuration reverses this ordering ($u_{\text{fused}}$ $0.757$, $\hat{\sigma}^2_{\text{best}}$ $0.560$, Table \ref{tab:auroc-separability}) despite competitive detection performance (Table \ref{tab:loc}). Following a similar diagnostic approach of He et al. \cite{he2019boundingboxregressionuncertainty}, we examine the $\hat{\sigma}^2_{\text{best}}$ distributions conditioned on localization outcome (Appendix \ref{app:spatial}): IR success and failure frames have similar $\hat{\sigma}^2_{\text{best}}$ distributions (mean 0.022 vs. 0.030), while RGB distributions are more clearly separated (mean 0.025 vs. 0.050). IR's Gaussian head assigns similar variance regardless of whether the localization succeeded or failed. A possible explanation is the lower spatial resolution of the thermal sensor $(640\times512 \text{ vs. } 1920\times1080)$: smaller targets occupy fewer pixels, potentially leaving the head with less signal to work with. However, a supplementary analysis stratified by target size (Appendix~\ref{app:spatial}) rules this out: the gap between IR and RGB remains constant across small, medium and large target groups. A more likely cause is the well-documented texture asymmetry between modalities: RGB provides rich texture and color detail, whereas thermal images primarily convey structural contours \cite{deng2021feanetfeatureenhancedattentionnetwork, sun2019rtfnet}. Since a Gaussian head learns to associate variation with localization error, the weaker texture signal in thermal images leaves it outcome-insensitive. Which specific thermal property drives this (contrast, edge sharpness or other factors) requires controlled ablation and is left for future work.


The weaker specificity of fusion operators reflects the same low-conflict structure identified in Section~\ref{sec:disc-pred}: under the vacuous miss-encoding, a non-detecting stream pulls the fused belief toward UAV on presence-dominated data, making the system less willing to predict absence. This is the correct safety-critical behavior under ignorance, but reduces specificity as a side effect. One caveat specific to calibration: on a presence-dominated benchmark, a model that is confidently correct most of the time will appear well-calibrated by construction. ECE scores should therefore be interpreted alongside the specificity results, which expose the minority-class behavior (absent frames) that overall ECE hides.

\subsection{Robustness}



The analysis further examines whether uncertainty-aware fusion degrades more gracefully than single-modality baselines across the evaluated attribute subsets. It does, leading Acc on all attribute subsets and holding its ECE calibration advantage on 8 of 10 subsets (Table \ref{tab:attribute_test}). The monotone decline in fused Acc across thermal-crossover tiers confirms TC severity drives the hardest residual failures.


The attribute results reveal two directions of cross-modal complementarity. IR dominates under LI (Acc 0.681 vs. RGB 0.603), as expected from a thermal sensor's independence from visible light. Yet IR's uncertainty signal degrades in this condition: its $\rho(u, IoU)$ turns positive (+0.283), meaning IR assigns lower uncertainty where localization is worse. RGB's $\rho$ remains correctly negative (-0.400) and fusion corrects this toward zero (-0.036) while retaining IR's localization advantage. IR rescues RGB on detection, RGB rescues IR on uncertainty quality. This complementarity is an averaging effect: as established in Section \ref{sec:disc-pred}, the three belief-fusion operators are numerically indistinguishable on this data.

The complementarity reverses under TC-Easy, the one attribute where RGB leads IR on Acc (0.688 vs 0.657). When the drone's thermal signature blends with its background but its visual appearance remains distinct, RGB carries localization. Across TC tiers the picture is not monotone for the unimodal streams: at TC-Mid IR recovers above RGB (0.612 vs. 0.576), suggesting the relationship between crossover severity and modality advantage is non-linear. At TC-Hard both streams collapse similarly (0.395 vs. 0.408) and fusion provides only marginal improvement (0.463). This is consistent with the near-zero conflict structure: when both streams fail together, fusion has no corrective signal. IR leads RGB on most other attributes, with the largest gap under OV (0.673 vs 0.550) and LR (0.598 vs. 0.517). The two conditions discussed above (LI and TC) are the clearest cases where the direction of complementarity is theoretically interpretable. For the remaining attributes (OV, LR, SV), IR leads without a mechanism that the data here isolates. Establishing the direction of complementarity for the other attributes would require controlled degradation rather than observational attribute subsets.


Beyond these two conditions, the one counter-trend is uncertainty-failure coupling: RGB yields the sharpest $\rho$ on 9 of 10 subsets, with fusion consistently weaker. We believe this is a scale effect: fusion operates at a compressed $\bar{u}$ relative to either unimodal stream, and since streams agree on nearly every frame ($C \approx 0$, Section~\ref{sec:localization}), the fused uncertainty collapses into a narrow band whose range restriction weakens the observed $\rho$ \cite{sackett2000range}. That fusion is simultaneously best-calibrated on most subsets confirms $u_{\text{fused}}$ is reliable but compressed.

\subsection{Computational Efficiency}

The analysis further evaluates whether the dual-stream architecture satisfies real-time constraints and whether the choice of fusion operator influences throughput. The cost of multimodal perception is essentially the cost of running a second stream: throughput roughly halves as parameters and GFLOPs double, yet every variant stays above the 25 FPS target. The choice of fusion rule adds negligible cost, meaning DBF can be adopted without overhead penalty even on data where its discounting step has no effect. 

\subsection{Detection-Only vs. Detect-Track}
\label{sec:discussion_dettrack}

A secondary outcome of the architecture decision in Section~\ref{sec:method-framework}, where the detection-only configuration is adopted as the main setup, is that it matches or exceeds the detect-track variant described in Appendix~\ref{app:det-track}. Although this comparison is not a primary focus of the study, it is relevant because detect-track collaboration constitutes the central mechanism of the EDTC system extended in this work. Two explanations are consistent with this outcome.

First, in EDTC the evidential head functions as a post-hoc verification module: after the corner head produces a bounding box, the head crops the predicted box's contents and attends against a template, directly observing whether the tracked region still resembles the target. In our proposed detect+track framework with SiamCAR \cite{guo2019siamcarsiamesefullyconvolutional}, the evidential head was instead integrated into SiamCAR's classification branch, producing per-position target/background scores aggregated by peak-picking. When the tracker drifts, the peak score remains high at the drifted location, whereas EDTC's verification head would see the drifted content directly and return low confidence. The near-zero collapse of $b_{fused}(UAV)$ across all TRACK configurations (Appendix \ref{app:separability-histograms}, Figures~\ref{fig:hist-ir}-\ref{fig:hist-dbf}, Subfigure~(d)), is consistent with this account, though a direct comparison between the two head designs is required to establish it conclusively. 

Second, the detect-track paradigm derives its value from recovering targets after genuine absence-reappearance events. This is evidenced by the EDTC paper's own ablation, where YOLO alone (0.392 Acc) outperforms YOLO + tracker without the evidential head (0.352 Acc) on AntiUAV600. The tracker propagates false positives until the evidential head suppresses them \cite{Zhu2023EDTC}. Anti-UAV is presence-dominated (UAV present in $\approx94-99\%$ of frames, median absence of 28 frames), leaving few absence-reappearance events for a tracker to benefit from. The argument does not require any deficiency in SiamCAR specifically: even an optimal tracker would have limited room to improve over detection-only when absence segments are rare and brief. Disentangling these two factors would require either an implementation mirroring EDTC's verification-head design, or evaluation on a dataset with longer and more frequent absence segments.

\subsection{Limitations}
\label{sec:limitations}

\textbf{Generalizability.} The null result for DBF is conditional on two compounding properties: the benchmark's near-universal UAV presence leaves both streams agreeing on nearly every frame, and the vacuous encoding of non-detecting streams forces conflict to zero on any frame where one stream fails to detect (Section~\ref{sec:disc-pred}). Whether DBF differentiates itself under stronger modality asymmetry or with objectness-informed background belief on weak detections remains untested. \textbf{Validity.} The comparison against EDTC and EDTC* is indicative only, as those are thermal-only detect-track systems. ECE should be interpreted alongside specificity results (Table~\ref{tab:belief}): on a presence-dominated benchmark, a confidently correct model appears well-calibrated by construction. Attribute-level complementarity claims cannot be isolated without controlled modality degradation. \textbf{Reliability.} The OC subset contains only $n{=}2$ test sequences. Detect-track results (Appendix~\ref{app:det-track}) are single-seed under a preliminary protocol not directly comparable to Table~\ref{tab:loc}. \textbf{Scalability.} Efficiency measurements are A100-specific. The dual-stream overhead may bottleneck on edge hardware.

\section{Conclusion}
\label{sec:conclusion}

Existing multimodal anti-UAV systems fuse RGB and thermal streams deterministically, leaving them unable to express doubt when streams disagree. This paper extended EDTC to RGB-Thermal perception via Discounted Belief Fusion and asked what impact uncertainty-aware fusion  has on predictive performance, calibration, robustness, and efficiency.

Fusion consistently outperforms either single stream in detection performance, achieving a test Acc of 0.670 compared with 0.604 for IR and 0.598 for RGB, and across all evaluated attribute subsets, with IR performing better under low illumination and RGB under thermal crossover. The fused uncertainty is well-calibrated, with an ECE of 0.057, although spatial variance provides stronger discrimination of localization failures (AUROC 0.739 vs.\ 0.626). The dual-stream architecture satisfies real-time constraints at 38--41 FPS, with no measurable throughput penalty associated with the choice of fusion operator. However, DBF, ABF, and naive averaging are indistinguishable across all evaluated metrics. This represents a structural finding rather than a deficiency of the operator: near-zero conflict follows jointly from the benchmark's presence-dominance and the vacuous encoding of non-detecting streams, both of which suppress disagreement by construction. Overall, uncertainty-aware multimodal fusion reliably improves localization and robustness over single-stream baselines, while distinguishing the benefits of conflict-aware operators requires benchmarks exhibiting genuine cross-modal disagreement.

\textbf{Future Work.} The most promising path forward is routing low detector objectness into background belief on detected frames, while leaving true non-detections vacuous. This would activate DBF's discounting on weak detections without reintroducing the veto behavior the vacuous encoding was designed to prevent.

\onecolumn

\appendix

\section{List of Abbreviations}

\begin{table}[h]
\centering
\small
\begin{tabular}{@{}ll@{}}
\toprule
\textbf{Abbreviation} & \textbf{Definition} \\
\midrule
\multicolumn{2}{@{}l}{\textit{Task \& Dataset}} \\
UAV   & Unmanned Aerial Vehicle \\
IR    & Infrared \\
RGB   & Red-Green-Blue \\
TC    & Thermal Crossover \\
FM    & Fast Motion \\
SV    & Scale Variation \\
LI    & Low Illumination \\
LR    & Low Resolution \\
OV    & Out-of-View \\
OC    & Occlusion \\
\midrule
\multicolumn{2}{@{}l}{\textit{Framework \& Methods}} \\
EDTC  & Evidential Detection and Tracking Collaboration \\
EDL   & Evidential Deep Learning \\
DBF   & Discounted Belief Fusion \\
ABF   & Averaging Belief Fusion \\
DMS   & Dynamic Modality Selection \\
DST   & Dempster-Shafer Theory \\
NMS   & Non-Maximum Suppression \\
\midrule
\multicolumn{2}{@{}l}{\textit{Evaluation Metrics}} \\
IoU   & Intersection over Union \\
mIoU  & Mean Intersection over Union \\
AP    & Average Precision \\
mAP   & Mean Average Precision \\
Acc   & EDTC Accuracy metric \\
ECE   & Expected Calibration Error \\
AUROC & Area Under the Receiver Operating Characteristic curve \\
AUSE  & Area Under the Sparsification Error curve \\
FPS   & Frames Per Second \\
GFLOPs & Giga Floating-Point Operations per Second \\
\midrule
\multicolumn{2}{@{}l}{\textit{Technical Terms}} \\
ReLU  & Rectified Linear Unit \\
KL    & Kullback-Leibler (divergence) \\
NLL   & Negative Log-Likelihood \\
MLP   & Multi-Layer Perceptron \\
SOTA  & State of the Art \\
\bottomrule
\end{tabular}
\caption{List of abbreviations used throughout this paper.}
\label{tab:abbreviations}
\end{table}

\section{Spatial Uncertainty Discriminability Analysis}
\label{app:spatial}

This appendix investigates why IR's spatial variance $\hat{\sigma}^2_{\text{best}}$ is a near-chance failure detector (AUROC 0.560) while RGB's is strong (AUROC 0.748), despite IR being competitive on localization metrics (Table \ref{tab:loc}). The analysis uses detected present frames from the validation set, pooled across seeds 0–2. A frame counts as a success if $IoU \geq 0.5$ and as a failure if $0 < IoU < 0.5$. Frames where no detection was made are excluded.

\begin{figure}[htbp]
    \centering
    \includegraphics[width=0.7\linewidth]{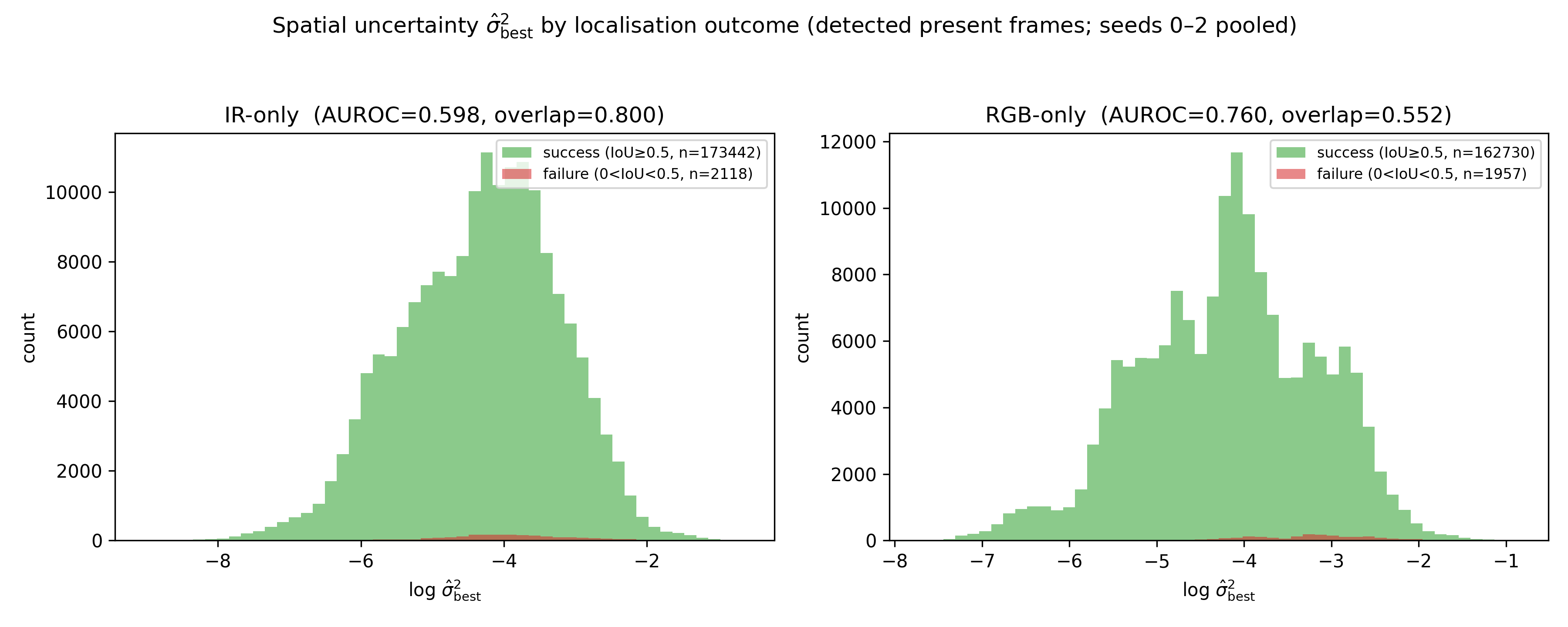}
    \caption{Distribution of log $\hat{\sigma}^2_{\text{best}}$ conditioned on localization outcome (success: $IoU \geq 0.5$, failure: $0 < IoU < 0.5$) for IR-only (left) and RGB-only (right) configurations, pooled across seeds 0–2 on the validation set. IR shows heavily overlapping distributions (overlap coefficient 0.800, AUROC 0.598), while RGB shows clearly separated distributions (overlap 0.552, AUROC 0.760). The contrast confirms that IR's Gaussian head produces outcome-insensitive variance, explaining the reversal of the spatial vs. semantic uncertainty ordering observed in Table \ref{tab:auroc-separability}.}
    \label{fig:distr}
\end{figure}

\textbf{Distribution analysis} Figure \ref{fig:distr} shows the distribution of $\log\hat{\sigma}^2_{\text{best}}$ separately for success and failure frames, for IR (left) and RGB (right). For RGB, failure frames have clearly higher variance than success frames (mean 0.050 vs. 0.025). The head assigns more uncertainty when it localizes poorly. For IR, the two distributions are nearly on top of each other (mean 0.030 vs. 0.022). The head assigns similar variance regardless of whether the localization was good or bad. The overlap coefficient (OVL) in the figure title quantifies this: it measures the fraction of shared area between the two distributions, where 1.0 means completely indistinguishable and 0.0 means perfectly separated. IR's OVL of 0.800 confirms the distributions are nearly identical. RGB's OVL of 0.552 confirms they are meaningfully separated.

\textbf{Size stratification analysis} A natural concern is that IR's outcome-insensitive variance is simply because IR frames have lower spatial resolution ($640\times512 vs. 1920
\times1080$), leaving fewer pixels per target for the Gaussian head to work with. To test this, Table \ref{tab:size-strat} stratifies frames into three equal-sized groups by target size (small, medium, large) and reports the overlap coefficient (OVL) and AUROC per group, separately for IR and RGB. If resolution were the cause, the gap between IR and RGB would shrink for larger targets where IR has more pixels. Instead, the gap stays constant across all three size groups (OVL difference approximately +0.29 for each group). IR's outcome-insensitive variance is therefore not explained by target size or sensor resolution. What specific property of thermal imagery is responsible remains an open question.

\begin{table}[H]
  \centering
  \small
  \setlength{\tabcolsep}{5pt}
  \caption{Target-size-stratified discriminability of the spatial uncertainty
    $\hat{\sigma}^2_{\text{best}}$ for the IR-only and RGB-only detectors,
    on detected present frames (IoU\,$>$\,0; success: IoU\,$\geq$\,0.5,
    failure: $0<$\,IoU\,$<$\,0.5), validation set, seeds 0--2 pooled.
    Frames are divided into three equal-sized groups by normalized predicted target size     $(w\!\cdot\! h)/A_{\text{frame}}$, with group boundaries computed per modality.
    OVL is the histogram overlap coefficient of $\log\hat{\sigma}^2_{\text{best}}$
    between the success and failure distributions (1\,=\,indistinguishable,
    0\,=\,disjoint); AUROC scores $\hat{\sigma}^2_{\text{best}}$ as a failure
    predictor. The IR$-$RGB gap is constant across sizes, isolating the
    thermal modality rather than target size as the cause.}
  \label{tab:size-strat}
  \begin{tabular}{llccccccc}
    \toprule
    Modality & Bin & Size range & $n_{\text{succ}}$ & $n_{\text{fail}}$
      & $\bar{\sigma}^2_{\text{succ}}$ & $\bar{\sigma}^2_{\text{fail}}$
      & OVL & AUROC \\
    & & ($\times10^{-3}$) & & & & & & \\
    \midrule
    \multirow{3}{*}{IR}
      & small  & $0.73$--$3.81$  & 57476 & 1076 & 0.023 & 0.033 & 0.836 & 0.606 \\
      & medium & $3.81$--$5.40$  & 58111 &  526 & 0.023 & 0.029 & 0.821 & 0.580 \\
      & large  & $5.40$--$30.4$  & 57855 &  516 & 0.022 & 0.028 & 0.729 & 0.619 \\
    \midrule
    \multirow{3}{*}{RGB}
      & small  & $0.23$--$2.52$  & 53663 & 1272 & 0.031 & 0.059 & 0.543 & 0.752 \\
      & medium & $2.52$--$4.74$  & 54488 &  452 & 0.022 & 0.026 & 0.527 & 0.647 \\
      & large  & $4.74$--$26.1$  & 54579 &  233 & 0.020 & 0.041 & 0.465 & 0.824 \\
    \bottomrule
  \end{tabular}
\end{table}

\section{Full Detect-Track Architecture}
\label{app:det-track}

\subsection{Methodology}
\begin{figure}[H]
    \centering
    \includegraphics[width=0.5\linewidth]{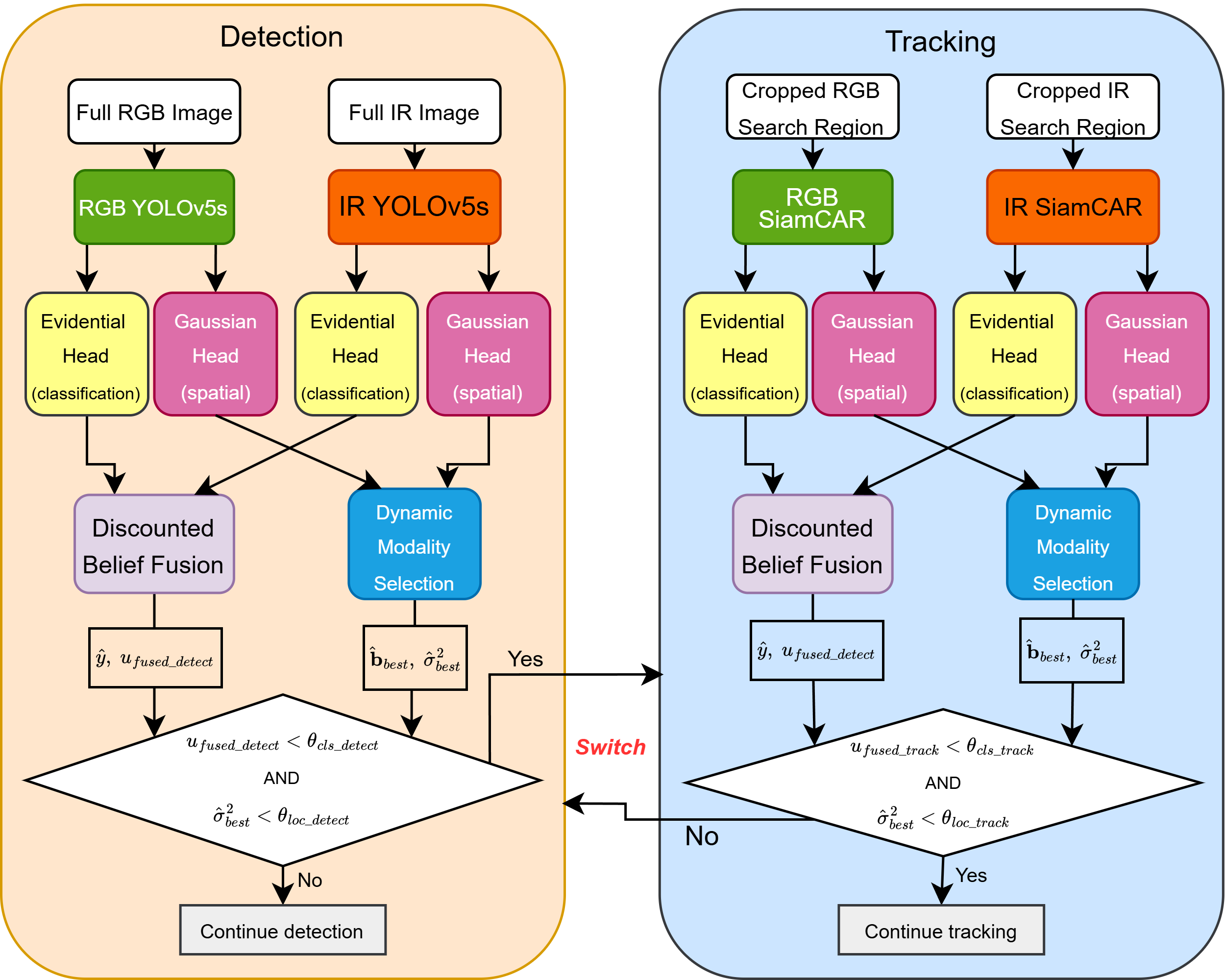}
    \caption{Multimodal uncertainty-aware detection-tracking framework. Both the global detection branch (YOLOv5s) and local tracking branch (SiamCAR) process RGB and infrared streams independently, each producing per-modality semantic uncertainty via an evidential head and spatial uncertainty via a Gaussian head. Semantic predictions are aggregated across modalities via Discounted Belief Fusion (DBF). Spatial predictions are resolved by selecting the bounding box from the lower-uncertainty modality (Dynamic Modality Selection). State-dependent switching between detection and tracking requires both semantic ($u_\text{fused}$) and spatial ($\hat{\sigma}^2_\text{best}$) confidence to satisfy their respective thresholds simultaneously.}
    \label{fig:architecture-full}
\end{figure}

The detect-track configuration follows the same dual-stream evidential pipeline as the main framework (Section~\ref{sec:methodology}), with three differences. Figure~\ref{fig:architecture-full} provides a high-level overview of the architecture.

\textbf{Architecture.} A local tracking branch is added alongside the global detection branch. EDTC's transformer-based tracker is replaced by SiamCAR \cite{guo2019siamcarsiamesefullyconvolutional}, an anchor-free Siamese tracker with a ResNet-50 backbone, equipped with the same evidential and Gaussian heads as the detection branch. The evidential head is integrated into SiamCAR's classification branch; the frame-level opinion is read at the peak position of the resulting score map.

\textbf{Tracking training.} Each modality's SiamCAR tracker is initialised from a checkpoint pretrained on VID, YOUTUBEBB, DET, COCO, GOT-10k, and LaSOT, then fine-tuned on the Anti-UAV training split using standard SiamCAR training ($\lambda_1{=}1$, $\lambda_2{=}3$) for 20 epochs. The evidential framework is added in the same two-stage protocol as detection: Stage~1 trains Gaussian heads for 30 epochs (SGD, lr $10^{-4}$, weight decay $10^{-4}$, $\lambda_\text{var}{=}0.1$); Stage~2 trains the evidential head for 40 epochs (lr $10^{-4}$, cosine-annealed, $\lambda_\text{evi}{=}1.0$), with KL regularisation annealed from 0 to 1 over the first 10 epochs.

\textbf{Switching and threshold calibration.} The uncertainty-gated output decision is replaced by a full state-transition mechanism: the system initializes tracking on a confident detection and reverts to detection when tracking confidence falls. This requires four thresholds rather than one: $\theta_{\text{cls\_det}}$ and $\theta_{\text{loc\_det}}$ govern the detection-to-tracking transition; $\theta_{\text{cls\_trk}}$ and $\theta_{\text{loc\_trk}}$ govern tracking maintenance. All four require $b_{fused}(\text{UAV}) > b_{fused}(\text{bg})$ as an additional class condition. Because joint grid search over four thresholds is computationally intractable, thresholds are calibrated from a single inference pass: the 95th percentile of $u_\text{fused}$ and $\hat{\sigma}^2_\text{best}$ on frames where tracking was successfully initiated or maintained yields one threshold per signal per branch state, confirmed by a $\pm 0.1$ sensitivity analysis around each derived value.

\subsection{Results \& Discussion}
\subsubsection{Localization Performance}
The results in this appendix were obtained under a preliminary protocol  that differs from the main setup in two respects. First, the frame-level YOLO opinion was computed by mean aggregation over all anchors rather than at the maximum-confidence NMS survivor. Second, non-detecting frames were encoded as ($b_{bg}{=}1$, $u{=}1$), which violates the Subjective Logic constraint $\sum_k b_k + u = 1$ and produces undefined behavior in the fusion equations; the final framework uses the vacuous opinion ($b_{UAV}{=}b_{bg}{=}0$, $u{=}1$) instead. Both deviations disproportionately affect the detect-track branch, so the qualitative finding that detection-only matches or exceeds detect-track, is unlikely to reverse under the corrected protocol. Under this preliminary protocol, the DBF detect-track variant (Acc 0.797) nominally exceeds IR detector-only (Acc 0.771). This reflects the cross-modality comparison rather than a genuine detect-track advantage, since within the same modality DBF detector-only (Acc 0.802) still exceeds DBF detect-track. The absolute numbers are not directly comparable to Table~\ref{tab:loc} and should be read as indicative only.

\begin{table}[h]
\centering
\caption{Best-run Acc per method under the preliminary detect-track protocol (validation set, seed 0). Each row is the best threshold configuration from a sequential calibration sweep; intermediate runs are not shown. The detector-only rows use the same protocol for direct comparability. Neither group is comparable to Table~\ref{tab:loc}, which uses the corrected protocol.}
\label{tab:det-track-summary}
\begin{tabular}{llc}
\toprule
Configuration & Method & Acc \\
\midrule
\multirow{3}{*}{Detect-track} & IR only  & 0.756 \\
                              & RGB only & 0.714 \\
                              & DBF      & 0.797 \\
\midrule
\multirow{3}{*}{Detector-only} & IR only  & 0.771 \\
                               & RGB only & 0.811 \\
                               & DBF      & 0.802 \\
\bottomrule
\end{tabular}
\end{table}

\subsubsection{Switching Signal Separability}
\label{app:separability-histograms}

\begin{figure}[htbp]
    \centering

    \begin{subfigure}[t]{0.32\textwidth}
        \centering
        \includegraphics[width=\textwidth]{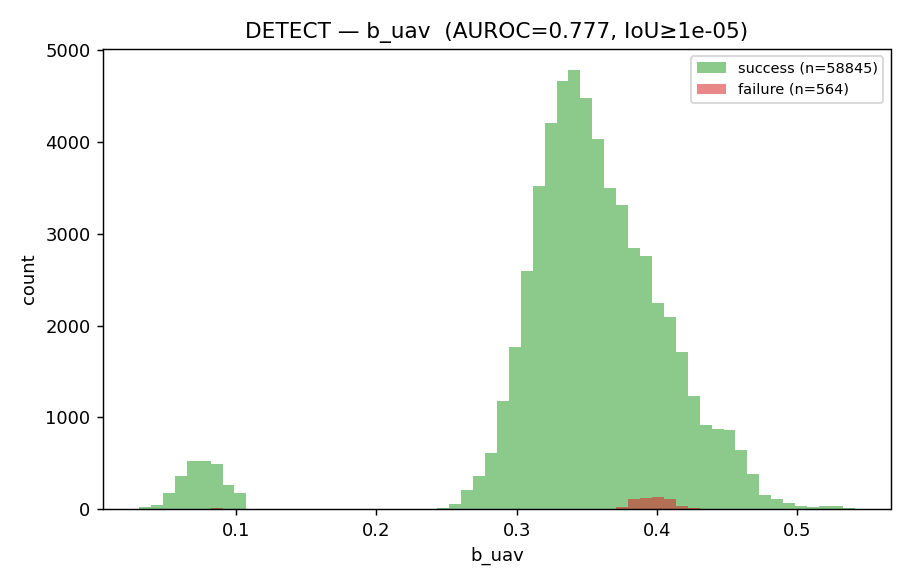}
        \caption{DETECT -- $b_\text{uav}$ (AUROC = 0.660)}
        \label{fig:ir-b-uav-detect}
    \end{subfigure}
    \hfill
    \begin{subfigure}[t]{0.32\textwidth}
        \centering
        \includegraphics[width=\textwidth]{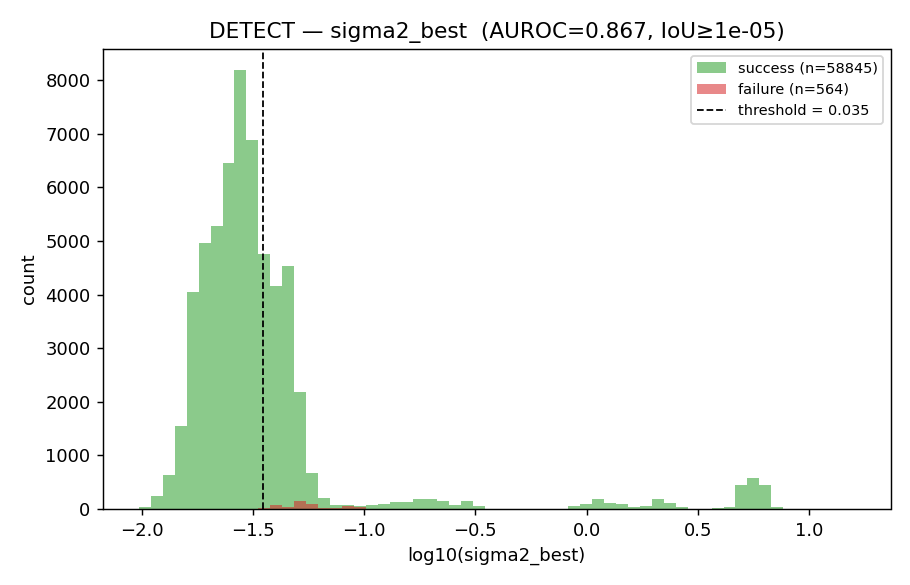}
        \caption{DETECT -- $\sigma^2_\text{best}$ (AUROC = 0.665)}
        \label{fig:ir-sigma2-detect}
    \end{subfigure}
    \hfill
    \begin{subfigure}[t]{0.32\textwidth}
        \centering
        \includegraphics[width=\textwidth]{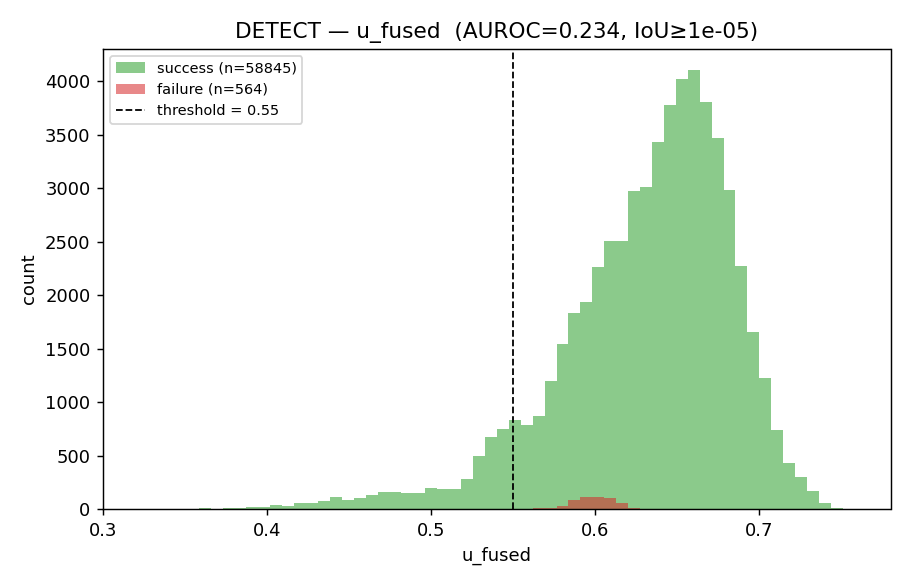}
        \caption{DETECT -- $u_\text{fused}$ (AUROC = 0.341)}
        \label{fig:ir-u-fused-detect}
    \end{subfigure}

    \vspace{0.5em}

    \begin{subfigure}[t]{0.32\textwidth}
        \centering
        \includegraphics[width=\textwidth]{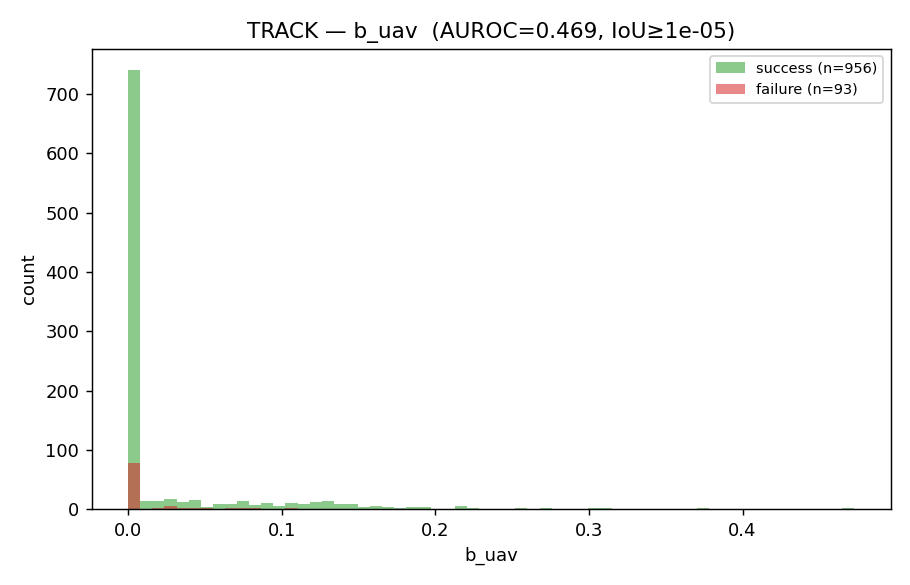}
        \caption{TRACK -- $b_\text{uav}$ (AUROC = 0.512)}
        \label{fig:ir-b-uav-track}
    \end{subfigure}
    \hfill
    \begin{subfigure}[t]{0.32\textwidth}
        \centering
        \includegraphics[width=\textwidth]{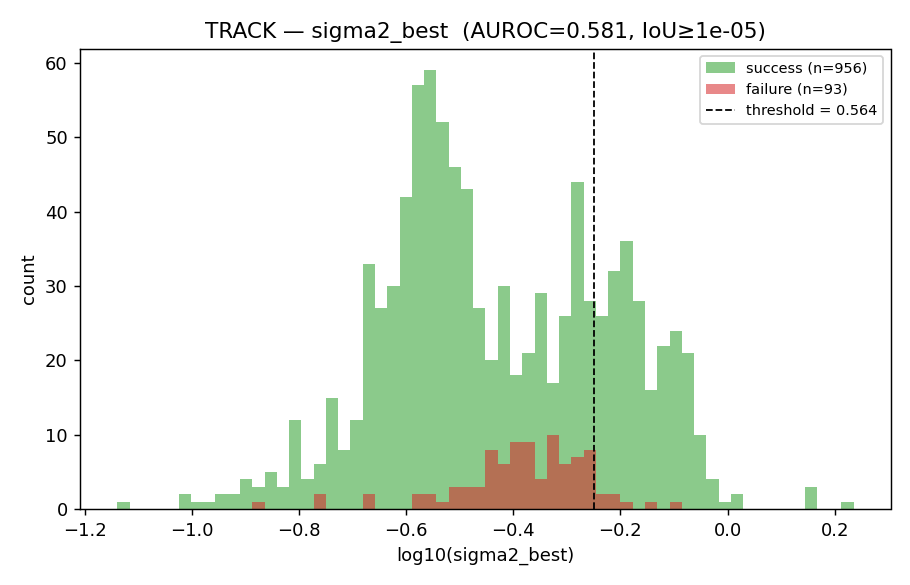}
        \caption{TRACK -- $\sigma^2_\text{best}$ (AUROC = 0.703)}
        \label{fig:ir-sigma2-track}
    \end{subfigure}
    \hfill
    \begin{subfigure}[t]{0.32\textwidth}
        \centering
        \includegraphics[width=\textwidth]{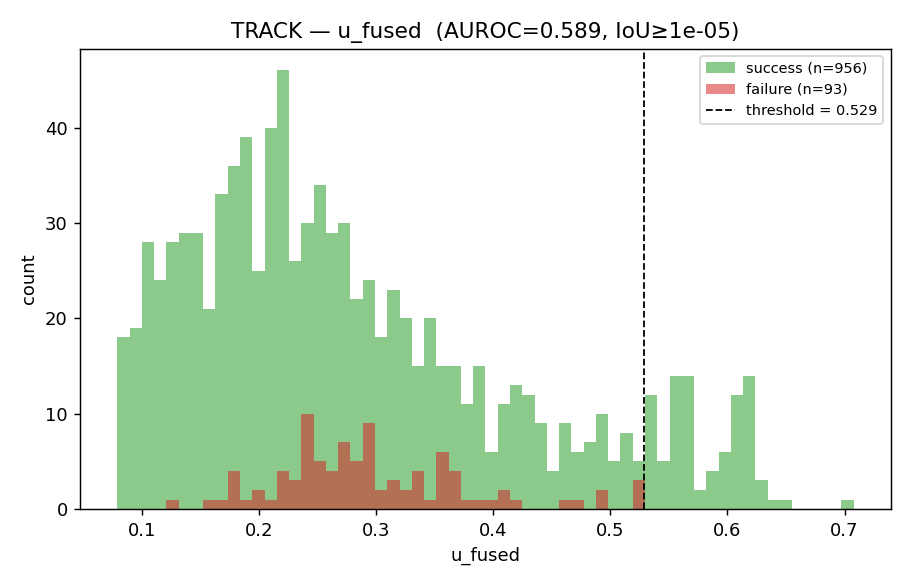}
        \caption{TRACK -- $u_\text{fused}$ (AUROC = 0.492)}
        \label{fig:ir-u-fused-track}
    \end{subfigure}

    \caption{IR-only configuration. Histograms of the three candidate switching signals across DETECT (top row) and TRACK (bottom row) modes on the validation set. Each subcaption reports the AUROC of the signal as a failure predictor.}
    \label{fig:hist-ir}
\end{figure}

\begin{figure}[htbp]
    \centering

    \begin{subfigure}[t]{0.32\textwidth}
        \centering
        \includegraphics[width=\textwidth]{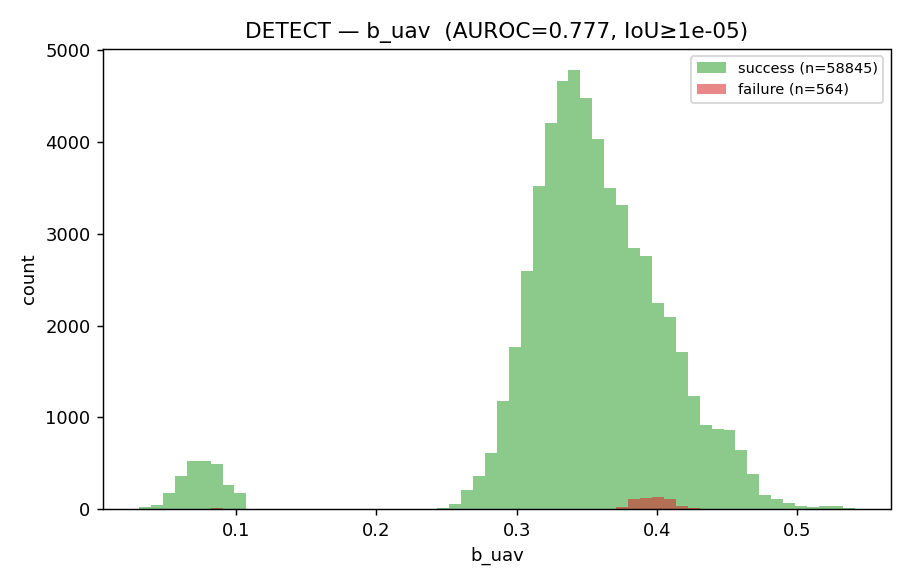}
        \caption{DETECT -- $b_\text{uav}$ (AUROC = 0.846)}
        \label{fig:rgb-b-uav-detect}
    \end{subfigure}
    \hfill
    \begin{subfigure}[t]{0.32\textwidth}
        \centering
        \includegraphics[width=\textwidth]{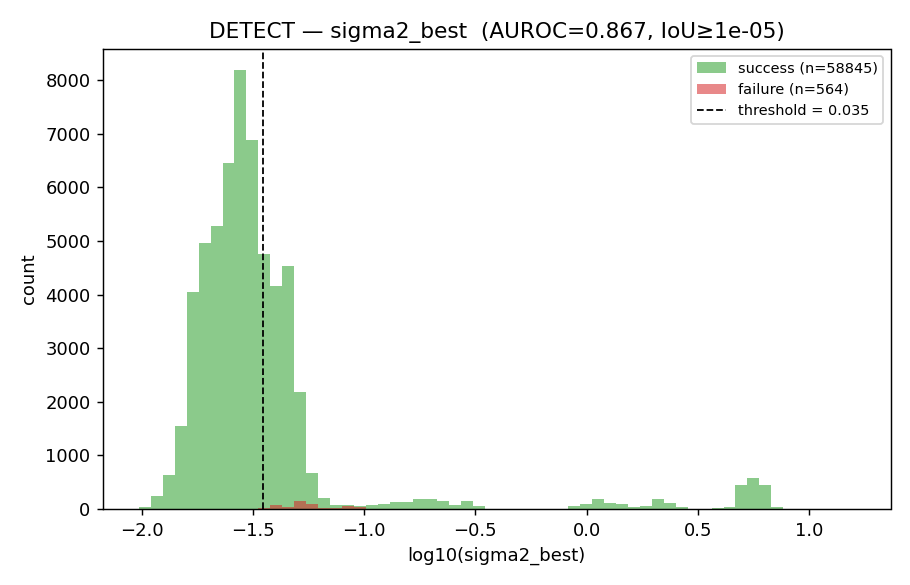}
        \caption{DETECT -- $\sigma^2_\text{best}$ (AUROC = 0.930)}
        \label{fig:rgb-sigma2-detect}
    \end{subfigure}
    \hfill
    \begin{subfigure}[t]{0.32\textwidth}
        \centering
        \includegraphics[width=\textwidth]{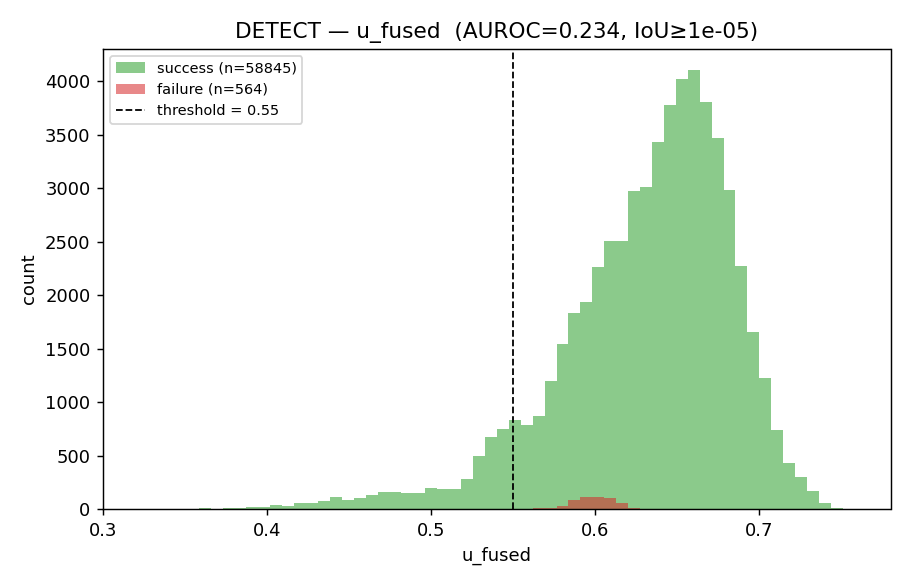}
        \caption{DETECT -- $u_\text{fused}$ (AUROC = 0.151)}
        \label{fig:rgb-u-fused-detect}
    \end{subfigure}

    \vspace{0.5em}

    \begin{subfigure}[t]{0.32\textwidth}
        \centering
        \includegraphics[width=\textwidth]{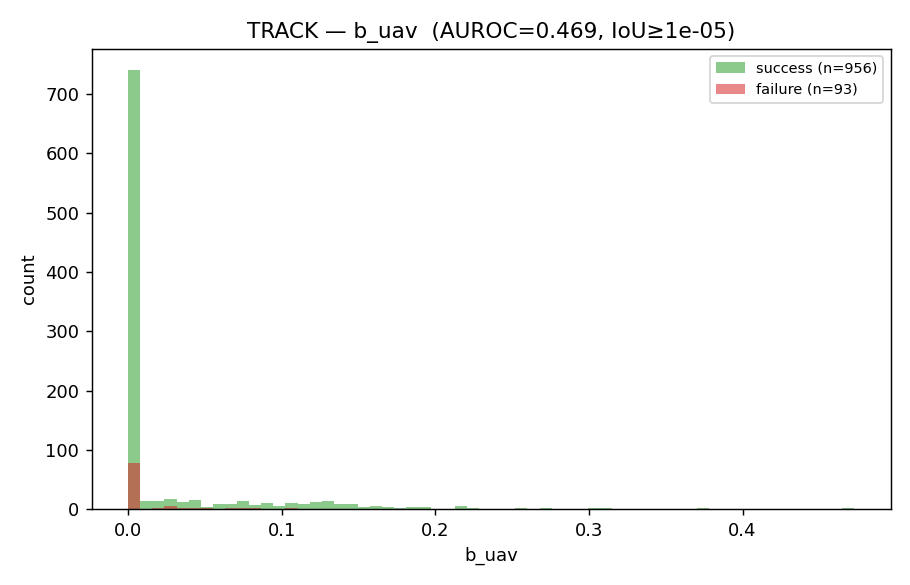}
        \caption{TRACK -- $b_\text{uav}$ (AUROC = 0.539)}
        \label{fig:rgb-b-uav-track}
    \end{subfigure}
    \hfill
    \begin{subfigure}[t]{0.32\textwidth}
        \centering
        \includegraphics[width=\textwidth]{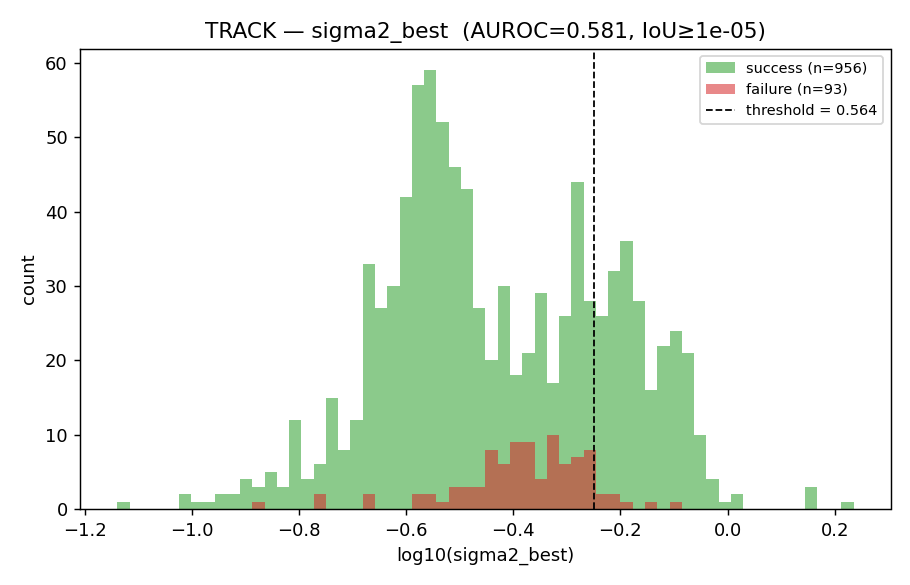}
        \caption{TRACK -- $\sigma^2_\text{best}$ (AUROC = 0.867)}
        \label{fig:rgb-sigma2-track}
    \end{subfigure}
    \hfill
    \begin{subfigure}[t]{0.32\textwidth}
        \centering
        \includegraphics[width=\textwidth]{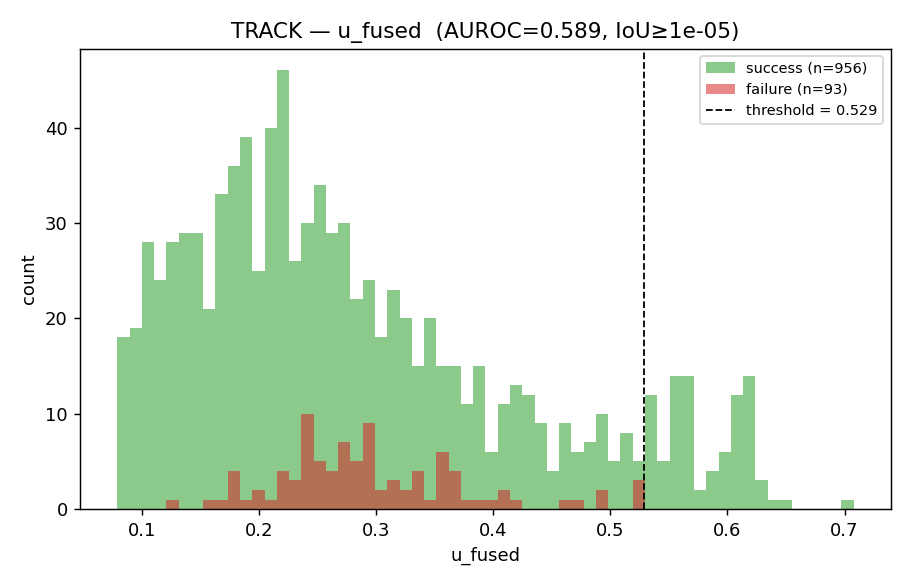}
        \caption{TRACK -- $u_\text{fused}$ (AUROC = 0.234)}
        \label{fig:rgb-u-fused-track}
    \end{subfigure}

    \caption{RGB-only configuration. Histograms of the three candidate switching signals across DETECT (top row) and TRACK (bottom row) modes on the validation set. Each subcaption reports the AUROC of the signal as a failure predictor.}
    \label{fig:hist-rgb}
\end{figure}

\begin{figure}[htbp]
    \centering

    \begin{subfigure}[t]{0.32\textwidth}
        \centering
        \includegraphics[width=\textwidth]{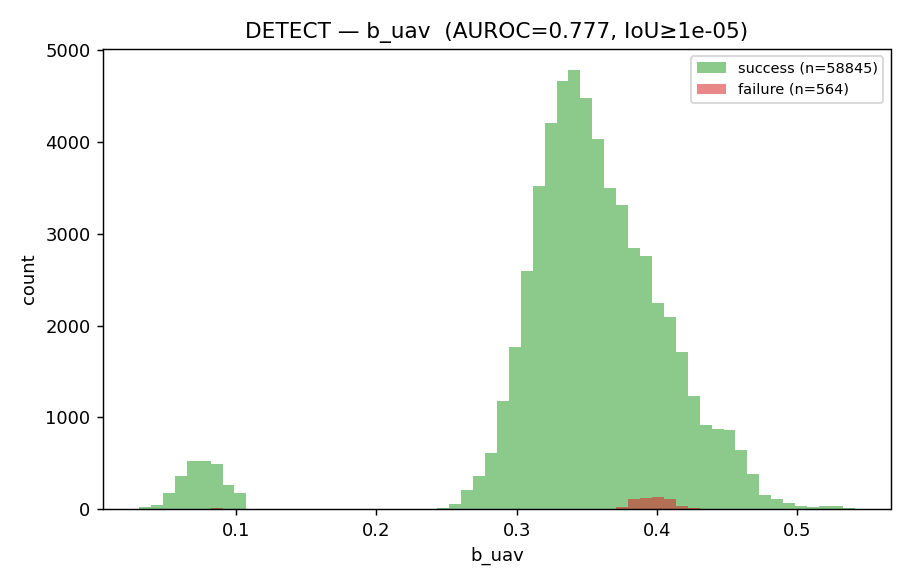}
        \caption{DETECT -- $b_\text{uav}$ (AUROC = 0.777)}
        \label{fig:dbf-b-uav-detect}
    \end{subfigure}
    \hfill
    \begin{subfigure}[t]{0.32\textwidth}
        \centering
        \includegraphics[width=\textwidth]{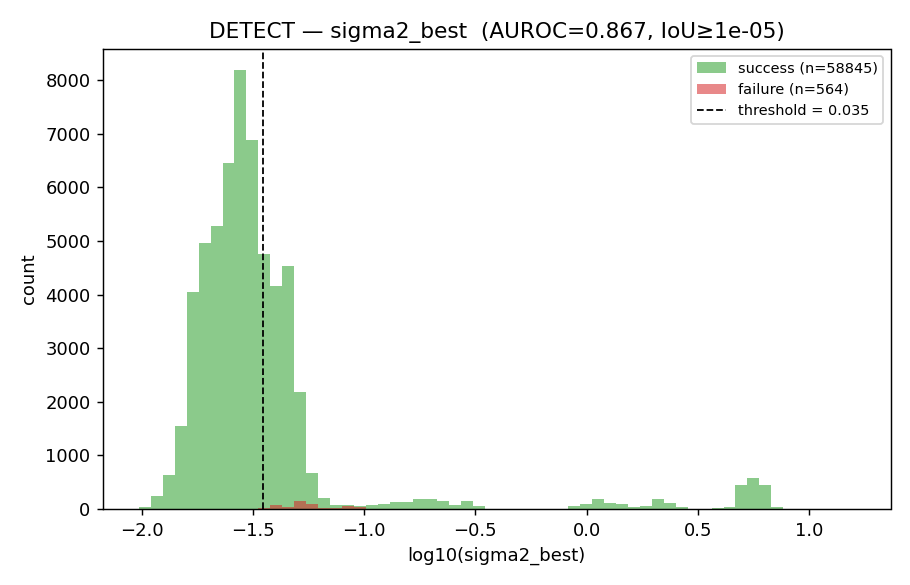}
        \caption{DETECT -- $\sigma^2_\text{best}$ (AUROC = 0.867)}
        \label{fig:dbf-sigma2-detect}
    \end{subfigure}
    \hfill
    \begin{subfigure}[t]{0.32\textwidth}
        \centering
        \includegraphics[width=\textwidth]{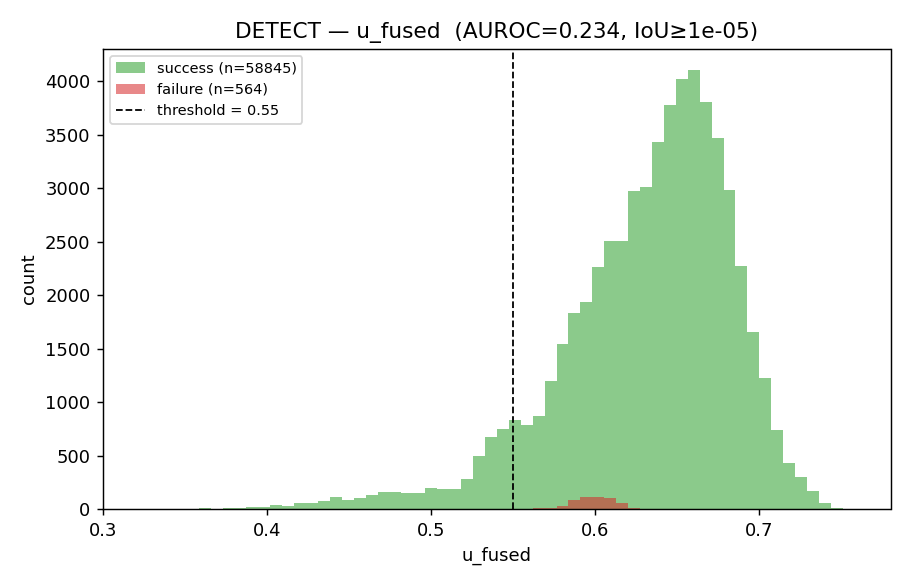}
        \caption{DETECT -- $u_\text{fused}$ (AUROC = 0.234)}
        \label{fig:dbf-u-fused-detect}
    \end{subfigure}

    \vspace{0.5em}

    \begin{subfigure}[t]{0.32\textwidth}
        \centering
        \includegraphics[width=\textwidth]{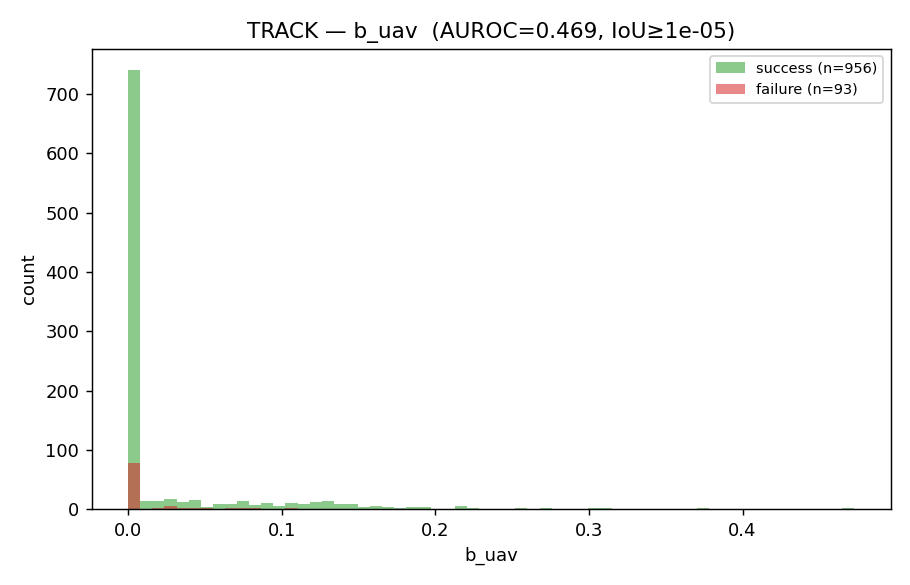}
        \caption{TRACK -- $b_\text{uav}$ (AUROC = 0.469)}
        \label{fig:dbf-b-uav-track}
    \end{subfigure}
    \hfill
    \begin{subfigure}[t]{0.32\textwidth}
        \centering
        \includegraphics[width=\textwidth]{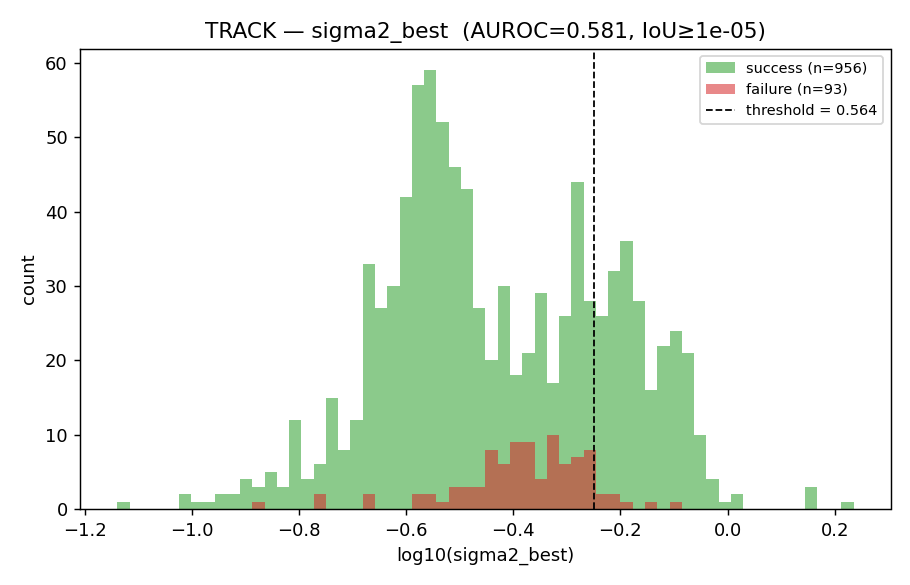}
        \caption{TRACK -- $\sigma^2_\text{best}$ (AUROC = 0.581)}
        \label{fig:dbf-sigma2-track}
    \end{subfigure}
    \hfill
    \begin{subfigure}[t]{0.32\textwidth}
        \centering
        \includegraphics[width=\textwidth]{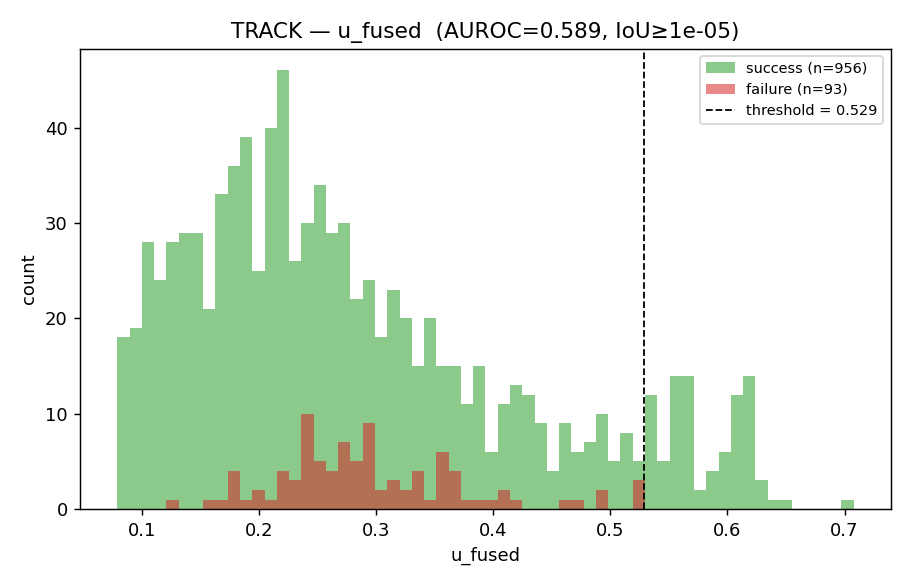}
        \caption{TRACK -- $u_\text{fused}$ (AUROC = 0.589)}
        \label{fig:dbf-u-fused-track}
    \end{subfigure}

    \caption{DBF configuration. Histograms of the three candidate switching signals across DETECT (top row) and TRACK (bottom row) modes on the validation set. Each subcaption reports the AUROC of the signal as a failure predictor.}
    \label{fig:hist-dbf}
\end{figure}

This appendix complements Section~\ref{sec:discussion_dettrack} by providing per-frame histograms of the three candidate switching signals ($b_{\text{fused}}(\text{UAV})$, $\hat{\sigma}^2_{\text{best}}$, $u_{\text{fused}}$) on the validation set. Figures are organised by modality: Figure~\ref{fig:hist-ir} (IR-only), Figure~\ref{fig:hist-rgb} (RGB-only), and Figure~\ref{fig:hist-dbf} (DBF). Each figure separates successes (IoU $\geq 10^{-5}$, green) from failures (IoU $=0$, red) and shows both switching states (DETECT, TRACK). Dashed lines indicate the operational thresholds used at evaluation.

The key pattern visible across all three modalities is the near-zero collapse of $b_{\text{fused}}(\text{UAV})$ in the TRACK rows, consistent with the drift argument discussed in Section \ref{sec:discussion_dettrack}. $\hat{\sigma}^2_{\text{best}}$ remains the more reliable failure signal across both states and all modalities.

\clearpage


\bibliographystyle{ACM-Reference-Format}
\bibliography{bibtex_acm_clean}

@ARTICLE{Jiang2023ANTIUAV,
  author={Jiang, Nan and Wang, Kuiran and Peng, Xiaoke and Yu, Xuehui and Wang, Qiang and Xing, Junliang and Li, Guorong and Guo, Guodong and Ye, Qixiang and Jiao, Jianbin and Zhao, Jian and Han, Zhenjun},
  journal={IEEE Transactions on Multimedia},
  title={Anti-UAV: A Large-Scale Benchmark for Vision-Based UAV Tracking},
  year={2023},
  volume={25},
  pages={486--500},
  doi={10.1109/TMM.2021.3128047}}

@misc{Zhu2023EDTC,
  author        = {Zhu, Xue-Feng and Xu, Tianyang and Zhao, Jian and Liu, Jia-Wei and Wang, Kai and Wang, Gang and Li, Jianan and Wang, Qiang and Jin, Lei and Zheng, Zhu and Xing, Junliang and Wu, Xiao-Jun},
  title         = {Evidential Detection and Tracking Collaboration: New Problem, Benchmark and Algorithm for Robust Anti-{UAV} System},
  year          = {2023},
  eprint        = {2306.15767},
  archivePrefix = {arXiv},
  primaryClass  = {cs.CV},
  doi           = {10.48550/arXiv.2306.15767},
  url           = {https://arxiv.org/abs/2306.15767}
}

@article{alla_trident_2025,
  author  = {Alla, Ildi and Yahia, Selma and Loscri, Valeria},
  title   = {{TRIDENT}: Tri-modal Real-time Intrusion Detection Engine for New Targets},
  journal = {Computers \& Security},
  volume  = {159},
  pages   = {104676},
  year    = {2025},
  doi     = {10.1016/j.cose.2025.104676}
}

@INPROCEEDINGS{YuanMMAUD,
  author={Yuan, Shenghai and Yang, Yizhuo and Nguyen, Thien Hoang and Nguyen, Thien-Minh and Yang, Jianfei and Liu, Fen and Li, Jianping and Wang, Han and Xie, Lihua},
  booktitle={2024 IEEE International Conference on Robotics and Automation (ICRA)},
  title={MMAUD: A Comprehensive Multi-Modal Anti-UAV Dataset for Modern Miniature Drone Threats},
  year={2024},
  pages={2745--2751},
  doi={10.1109/ICRA57147.2024.10610957}}

@inproceedings{dong2025securingskiescomprehensivesurvey,
  author    = {Dong, Yifei and Wu, Fengyi and Zhang, Sanjian and Chen, Guangyu and Hu, Yuzhi and Yano, Masumi and Sun, Jingdong and Huang, Siyu and Liu, Feng and Dai, Qi and Cheng, Zhi-Qi},
  title     = {Securing the Skies: A Comprehensive Survey on Anti-{UAV} Methods, Benchmarking, and Future Directions},
  booktitle = {Proceedings of the IEEE/CVF Conference on Computer Vision and Pattern Recognition Workshops (CVPRW)},
  pages     = {6725--6739},
  year      = {2025},
  doi       = {10.1109/CVPRW67362.2025.00663}
}

@article{JIAO20241,
title = {A Comprehensive Survey on Deep Learning Multi-Modal Fusion: Methods, Technologies and Applications},
journal = {Computers, Materials \& Continua},
volume = {80},
number = {1},
pages = {1--35},
year = {2024},
doi = {10.32604/cmc.2024.053204},
url = {https://www.sciencedirect.com/science/article/pii/S1546221824005216},
author = {Tianzhe Jiao and Chaopeng Guo and Xiaoyue Feng and Yuming Chen and Jie Song}
}

@article{Gao2025EDL,
  author  = {Gao, Junyu and Chen, Mengyuan and Xiang, Liangyu and Xu, Changsheng},
  title   = {A Comprehensive Survey on Evidential Deep Learning and Its Applications},
  journal = {IEEE Transactions on Pattern Analysis and Machine Intelligence},
  volume  = {48},
  number  = {3},
  pages   = {2118--2138},
  year    = {2026},
  doi     = {10.1109/TPAMI.2025.3625258}
}

@inproceedings{sensoy2018evidentialdeeplearningquantify,
  author    = {Sensoy, Murat and Kaplan, Lance and Kandemir, Melih},
  title     = {Evidential Deep Learning to Quantify Classification Uncertainty},
  booktitle = {Advances in Neural Information Processing Systems},
  volume    = {31},
  year      = {2018}
}

@article{baltrušaitis2017multimodalmachinelearningsurvey,
  author  = {Baltru{\v{s}}aitis, Tadas and Ahuja, Chaitanya and Morency, Louis-Philippe},
  title   = {Multimodal Machine Learning: A Survey and Taxonomy},
  journal = {IEEE Transactions on Pattern Analysis and Machine Intelligence},
  volume  = {41},
  number  = {2},
  pages   = {423--443},
  year    = {2019},
  doi     = {10.1109/TPAMI.2018.2798607}
}

@article{zhao2022visionbasedantiuavdetectiontracking,
  author  = {Zhao, Jie and Zhang, Jingshu and Li, Dongdong and Wang, Dong},
  title   = {Vision-Based Anti-{UAV} Detection and Tracking},
  journal = {IEEE Transactions on Intelligent Transportation Systems},
  volume  = {23},
  number  = {12},
  pages   = {25323--25334},
  year    = {2022},
  doi     = {10.1109/TITS.2022.3177627}
}

@misc{larrat2025multimodaltransformerapproachuav,
      title={A Multimodal Transformer Approach for UAV Detection and Aerial Object Recognition Using Radar, Audio, and Video Data},
      author={Mauro Larrat and Claudomiro Sales},
      year={2025},
      eprint={2511.15312},
      archivePrefix={arXiv},
      primaryClass={cs.CV},
      url={https://arxiv.org/abs/2511.15312}
}

@misc{zongzhen2025crossmodaloffsetguideddynamicalignment,
      title={Cross-modal Offset-guided Dynamic Alignment and Fusion for Weakly Aligned UAV Object Detection},
      author={Liu, Zongzhen and Luo, Hui and Wang, Zhixing and Wei, Yuxing and Zuo, Haorui and Zhang, Jianlin},
      year={2025},
      eprint={2506.16737},
      archivePrefix={arXiv},
      primaryClass={cs.CV},
      url={https://arxiv.org/abs/2506.16737}
}

@misc{fusionsubjectivelogic,
  author        = {van der Heijden, Rens W. and Kopp, Henning and Kargl, Frank},
  title         = {Multi-Source Fusion Operations in Subjective Logic},
  year          = {2018},
  eprint        = {1805.01388},
  archivePrefix = {arXiv},
  primaryClass  = {cs.AI},
  url           = {https://arxiv.org/abs/1805.01388}
}

@inproceedings{gal2016dropoutbayesianapproximationrepresenting,
  author    = {Gal, Yarin and Ghahramani, Zoubin},
  title     = {Dropout as a Bayesian Approximation: Representing Model Uncertainty in Deep Learning},
  booktitle = {Proceedings of the 33rd International Conference on Machine Learning},
  series    = {Proceedings of Machine Learning Research},
  volume    = {48},
  pages     = {1050--1059},
  year      = {2016},
  publisher = {PMLR},
  url       = {https://proceedings.mlr.press/v48/gal16.html}
}

@article{dempster,
author = {Srivastava, Rajendra},
year = {2022},
month = {08},
title = {Dempster-Shafer Theory of Belief Functions: A Language for Managing Uncertainties in the Real-World Problems},
journal = {International Journal of Finance, Entrepreneurship \& Sustainability},
doi = {10.56763/ijfes.v1i.30}
}

@inproceedings{guo2017calibrationmodernneuralnetworks,
  author    = {Guo, Chuan and Pleiss, Geoff and Sun, Yu and Weinberger, Kilian Q.},
  title     = {On Calibration of Modern Neural Networks},
  booktitle = {Proceedings of the 34th International Conference on Machine Learning},
  series    = {Proceedings of Machine Learning Research},
  volume    = {70},
  pages     = {1321--1330},
  year      = {2017},
  publisher = {PMLR},
  url       = {https://proceedings.mlr.press/v70/guo17a.html}
}

@inproceedings{geifman2017selectiveclassificationdeepneural,
  author    = {Geifman, Yonatan and El-Yaniv, Ran},
  title     = {Selective Classification for Deep Neural Networks},
  booktitle = {Advances in Neural Information Processing Systems},
  volume    = {30},
  year      = {2017}
}

@inproceedings{ilg2018uncertaintyestimatesmultihypothesesnetworks,
  author    = {Ilg, Eddy and {\c{C}}i{\c{c}}ek, {\"O}zg{\"u}n and Galesso, Silvio and Klein, Aaron and Makansi, Osama and Hutter, Frank and Brox, Thomas},
  title     = {Uncertainty Estimates and Multi-Hypotheses Networks for Optical Flow},
  booktitle = {Proceedings of the European Conference on Computer Vision (ECCV)},
  pages     = {652--667},
  year      = {2018}
}

@inproceedings{reddi2020mlperfinferencebenchmark,
  author    = {Reddi, Vijay Janapa and Cheng, Christine and Kanter, David and Mattson, Peter and Schmuelling, Guenther and Wu, Carole-Jean and Anderson, Brian and Breughe, Maximilien and Charlebois, Mark and Chou, William and Chukka, Ramesh and Coleman, Cody and Davis, Sam and Deng, Pan and Diamos, Greg and Duke, Jared and Fick, Dave and Gardner, J. Scott and Hubara, Itay and Idgunji, Sachin and Jablin, Thomas B. and Jiao, Jeff and St. John, Tom and Kanwar, Pankaj and Lee, David and Liao, Jeffery and Lokhmotov, Anton and Massa, Francisco and Meng, Peng and Micikevicius, Paulius and Osborne, Colin and Pekhimenko, Gennady and Rajan, Arun Tejusve Raghunath and Sequeira, Dilip and Sirasao, Ashish and Sun, Fei and Tang, Hanlin and Thomson, Michael and Wei, Frank and Wu, Ephrem and Xu, Lingjie and Yamada, Koichi and Yu, Bing and Yuan, George and Zhong, Aaron and Zhang, Peizhao and Zhou, Yuchen},
  title     = {{MLPerf} Inference Benchmark},
  booktitle = {2020 ACM/IEEE 47th Annual International Symposium on Computer Architecture (ISCA)},
  pages     = {446--459},
  year      = {2020},
  doi       = {10.1109/ISCA45697.2020.00045}
}

@INPROCEEDINGS{Josang2017SL,
  author={J{\o}sang, Audun and Wang, Dongxia and Zhang, Jie},
  booktitle={2017 20th International Conference on Information Fusion (Fusion)},
  title={Multi-source fusion in subjective logic},
  year={2017},
  pages={1--8},
  doi={10.23919/ICIF.2017.8009820}}

@inproceedings{bezirganyan2025multimodallearninguncertaintyquantification,
  author    = {Bezirganyan, Grigor and Sellami, Sana and Berti-{\'E}quille, Laure and Fournier, S{\'e}bastien},
  title     = {Multimodal Learning with Uncertainty Quantification Based on Discounted Belief Fusion},
  booktitle = {Proceedings of the 28th International Conference on Artificial Intelligence and Statistics (AISTATS)},
  series    = {Proceedings of Machine Learning Research},
  volume    = {258},
  pages     = {3142--3150},
  year      = {2025},
  publisher = {PMLR},
  url       = {https://proceedings.mlr.press/v258/bezirganyan25a.html}
}

@inproceedings{hendrycks2018baselinedetectingmisclassifiedoutofdistribution,
  author    = {Hendrycks, Dan and Gimpel, Kevin},
  title     = {A Baseline for Detecting Misclassified and Out-of-Distribution Examples in Neural Networks},
  booktitle = {International Conference on Learning Representations (ICLR)},
  year      = {2017},
  url       = {https://openreview.net/forum?id=Hkg4TI9xl}
}

@inproceedings{he2019boundingboxregressionuncertainty,
  author    = {He, Yihui and Zhu, Chenchen and Wang, Jianren and Savvides, Marios and Zhang, Xiangyu},
  title     = {Bounding Box Regression with Uncertainty for Accurate Object Detection},
  booktitle = {Proceedings of the IEEE/CVF Conference on Computer Vision and Pattern Recognition (CVPR)},
  pages     = {2888--2897},
  year      = {2019}
}

@inproceedings{lakshminarayanan2017simplescalablepredictiveuncertainty,
  author    = {Lakshminarayanan, Balaji and Pritzel, Alexander and Blundell, Charles},
  title     = {Simple and Scalable Predictive Uncertainty Estimation Using Deep Ensembles},
  booktitle = {Advances in Neural Information Processing Systems},
  volume    = {30},
  pages     = {6402--6413},
  year      = {2017}
}

@inproceedings{guo2019siamcarsiamesefullyconvolutional,
  author    = {Guo, Dongyan and Wang, Jun and Cui, Ying and Wang, Zhenhua and Chen, Shengyong},
  title     = {{SiamCAR}: Siamese Fully Convolutional Classification and Regression for Visual Tracking},
  booktitle = {Proceedings of the IEEE/CVF Conference on Computer Vision and Pattern Recognition (CVPR)},
  pages     = {6269--6277},
  year      = {2020}
}

@inproceedings{redmon2016lookonceunifiedrealtime,
  author    = {Redmon, Joseph and Divvala, Santosh and Girshick, Ross and Farhadi, Ali},
  title     = {You Only Look Once: Unified, Real-Time Object Detection},
  booktitle = {Proceedings of the IEEE Conference on Computer Vision and Pattern Recognition (CVPR)},
  pages     = {779--788},
  year      = {2016},
  doi       = {10.1109/CVPR.2016.91}
}

@article{sackett2000range,
author = {Sackett, Paul and Yang, Hyuckseung},
year = {2000},
month = {02},
pages = {112--118},
title = {Correction for Range Restriction: An Expanded Typology},
volume = {85},
journal = {Journal of Applied Psychology},
doi = {10.1037/0021-9010.85.1.112}
}

@inproceedings{deng2021feanetfeatureenhancedattentionnetwork,
  author    = {Deng, Fuqin and Feng, Hua and Liang, Mingjian and Wang, Hongmin and Yang, Yong and Gao, Yuan and Chen, Junfeng and Hu, Junjie and Guo, Xiyue and Lam, Tin Lun},
  title     = {{FEANet}: Feature-Enhanced Attention Network for {RGB}-Thermal Real-Time Semantic Segmentation},
  booktitle = {2021 IEEE/RSJ International Conference on Intelligent Robots and Systems (IROS)},
  pages     = {4467--4473},
  year      = {2021},
  doi       = {10.1109/IROS51168.2021.9636084}
}

@ARTICLE{sun2019rtfnet,
author={Yuxiang Sun and Weixun Zuo and Ming Liu},
journal={{IEEE Robotics and Automation Letters}},
title={{RTFNet: RGB-Thermal Fusion Network for Semantic Segmentation of Urban Scenes}},
year={2019},
volume={4},
number={3},
pages={2576--2583},
doi={10.1109/LRA.2019.2904733},
month={July},}

@inproceedings{AnwarZiabari2025,
  author    = {Anwar, Karim and Mohammadi Ziabari, Seyed Sahand},
  title     = {Attention to the Branches: A Comparative Analysis of FairMOT with Transformers on Fish Dataset},
  booktitle = {Multi-disciplinary Trends in Artificial Intelligence: 17th International Conference, MIWAI 2024, Pattaya, Thailand, November 11--15, 2024, Proceedings, Part I},
  series    = {Lecture Notes in Computer Science},
  volume    = {15431},
  pages     = {64--76},
  publisher = {Springer},
  year      = {2025},
  doi       = {10.1007/978-981-96-0692-4_6}
}

@inproceedings{Katona2025MARINE,
  author    = {Katona, Zs{\'o}fia and Mohammadi Ziabari, Seyed Sahand and Karimi Nejadasl, Fatemeh},
  title     = {{MARINE}: A Computer Vision Model for Detecting Rare Predator-Prey Interactions in Animal Videos},
  booktitle = {Big Data and Artificial Intelligence: 12th International Conference, BDA 2024, Hyderabad, India, December 17--20, 2024, Proceedings},
  series    = {Lecture Notes in Computer Science},
  volume    = {15526},
  pages     = {183--199},
  publisher = {Springer},
  year      = {2025},
  doi       = {10.1007/978-3-031-81821-9_11}
}

\end{document}